\documentclass[preprint,12pt]{elsarticle}
\usepackage[margin=2.5cm]{geometry}
\usepackage{xcolor}
\usepackage{booktabs}
\usepackage{setspace}
\usepackage{placeins}
\usepackage{booktabs}
\usepackage{adjustbox}
\usepackage{tabularx}
\usepackage{hyperref}
\usepackage{pdflscape}
\usepackage{caption}
\usepackage{longtable}
\usepackage{float}
\usepackage{soul}
\usepackage{setspace}
\usepackage{caption}
\usepackage{hypcap}
\usepackage{ragged2e}
\usepackage{array}
\usepackage[utf8]{inputenc}
\usepackage[T1]{fontenc}
\DeclareUnicodeCharacter{03BC}{\ensuremath{\mu}}
\usepackage{gensymb}

\hypersetup{
    colorlinks=true,
    linkcolor=blue,
    filecolor=magenta,      
    urlcolor=cyan
}
\usepackage{amssymb}
\usepackage{amsmath}
\usepackage{textcomp}
\begin{document}

\begin{frontmatter}

%% Title, authors and addresses

%% use the tnoteref command within \title for footnotes;
%% use the tnotetext command for theassociated footnote;
%% use the fnref command within \author or \address for footnotes;
%% use the fntext command for theassociated footnote;
%% use the corref command within \author for corresponding author footnotes;
%% use the cortext command for theassociated footnote;
%% use the ead command for the email address,
%% and the form \ead[url] for the home page:
%% \title{Title\tnoteref{label1}}
%% \tnotetext[label1]{}
%% \author{Name\corref{cor1}\fnref{label2}}
%% \ead{email address}
%% \ead[url]{home page}
%% \fntext[label2]{}
%% \cortext[cor1]{}
%% \affiliation{organization={},
%%             addressline={},
%%             city={},
%%             postcode={},
%%             state={},
%%             country={}}
%% \fntext[label3]{}
\title{NANOGPT: A Query-Driven Large Language Model Retrieval-Augmented Generation System for Nanotechnology Research}
% \title{Linking Two-Color Imaging of the Melt Pool to ex-situ Melt Depth Morphologies With Vision Transformers}
% Meltpool Depth Prediction Using Surface Thermal Images with Vision Transformers
%% use optional labels to link authors explicitly to addresses:
%% \author[label1,label2]{}
%% \affiliation[label1]{organization={},
%%             addressline={},
%%             city={},
%%             postcode={},
%%             state={},
%%             country={}}
%%
%% \affiliation[label2]{organization={},
%%             addressline={},
%%             city={},
%%             postcode={},
%%             state={},
%%             country={}}

% \author[inst1]{Achuth Chandrasekhar}

% \affiliation[inst1]{organization={Mechanical Engineering},%Department and Organization
%             addressline={Carnegie Mellon University}, 
%             city={Pittsburgh},
%             postcode={15213}, 
%             state={PA},
%             country={USA}}

% \author[inst1]{Jonathan Chan}
% \author[inst1]{Amir Barati Farimani}

\author[inst1]{Achuth Chandrasekhar}

\affiliation[inst1]{organization={Materials Science and Engineering},%Department and Organization
            addressline={Carnegie Mellon University}, 
            city={Pittsburgh},
            postcode={15213}, 
            state={PA},
            country={USA}}

\author[inst2]{Omid Barati Farimani}
\author[inst2]{Olabode T. Ajenifujah}
\author[inst4]{Janghoon Ock}
\author[inst2,inst3,inst4,inst5]{Amir Barati Farimani\corref{cor1}}
\ead{barati@cmu.edu}
\cortext[cor1]{Corresponding author}

\affiliation[inst2]{organization={Mechanical Engineering},%Department and Organization
            addressline={Carnegie Mellon University}, 
            city={Pittsburgh},
            postcode={15213}, 
            state={PA},
            country={USA}}

\affiliation[inst3]{organization={Biomedical Engineering},%Department and Organization
            addressline={Carnegie Mellon University}, 
            city={Pittsburgh},
            postcode={15213}, 
            state={PA},
            country={USA}}
\affiliation[inst4]{organization={Chemical Engineering},%Department and Organization
            addressline={Carnegie Mellon University}, 
            city={Pittsburgh},
            postcode={15213}, 
            state={PA},
            country={USA}}

\affiliation[inst5]{organization={Machine Learning Department},%Department and Organization
            addressline={Carnegie Mellon University}, 
            city={Pittsburgh},
            postcode={15213}, 
            state={PA},
            country={USA}}

% \affiliation[inst2]{organization={Department Two},%Department and Organization
%             addressline={Address Two}, 
%             city={City Two},
%             postcode={22222}, 
%             state={State Two},
%             country={Country Two}}

\begin{abstract}
%% Text of abstract
This paper presents the development and application of a Large Language Model Retrieval-Augmented Generation (LLM-RAG) system tailored for nanotechnology research. The system leverages the capabilities of a sophisticated language model to serve as an intelligent research assistant, enhancing the efficiency and comprehensiveness of literature reviews in the nanotechnology domain. Central to this LLM-RAG system is its advanced query backend retrieval mechanism, which integrates data from multiple reputable sources. The system retrieves relevant literature by utilizing Google Scholar's advanced search, and scraping open-access papers from Elsevier, Springer Nature, and ACS Publications. This multifaceted approach ensures a broad and diverse collection of up-to-date scholarly articles and papers. The proposed system demonstrates significant potential in aiding researchers by providing a streamlined, accurate, and exhaustive literature retrieval process, thereby accelerating research advancements in nanotechnology. The effectiveness of the LLM-RAG system is validated through rigorous testing, illustrating its capability to significantly reduce the time and effort required for comprehensive literature reviews, while maintaining high accuracy, query relevance and outperforming standard, publicly available LLMS.
\end{abstract}

\end{frontmatter}

\section{Introduction}
\label{sec:introduction}

The Transformer model is a groundbreaking architecture in the field of deep learning and natural language processing (NLP). It was introduced by a team of researchers at Google in a paper titled "Attention Is All You Need," published in 2017\cite{vaswani2017attention}. The introduction of the Transformer marked a significant shift from previous models, such as recurrent neural networks (RNNs) and convolutional neural networks (CNNs) \cite{lecun2015deep}, by relying entirely on self-attention mechanisms rather than recurrence or convolution.
In recent years, LLMs have emerged as a transformative technology in the field of NLP. These models are typically based on deep neural networks, particularly the Transformer architecture, and are characterized by their capacity to process and generate human-like text. By utilizing vast amounts of training data, LLMs can learn to understand and produce coherent language, making them invaluable for a wide range of applications such as machine translation, text summarization, and conversational agents \cite{mctear2022conversational}, \cite{yang2024automatic}, \cite{van2024adapted}, \cite{nllb2024scaling}.
The fundamental advancement of LLMs is the ability to capture complex linguistic patterns through extensive pre-training on diverse datasets, followed by fine-tuning on specific tasks. This paradigm allows LLMs to generalize across various contexts and exhibit remarkable fluency and versatility in text generation. Notable examples of such models include OpenAI's GPT (Generative Pre-trained Transformer) series \cite{alto2023modern} and Google's BERT \cite{devlin2018bert}(Bidirectional Encoder Representations from Transformers), which have set new benchmarks across multiple NLP tasks.

LLMs can be categorized into three main types: encoder-only models, decoder-only models, and encoder-decoder models. Encoder-only models excel at classifying sequences of text and are mainly used for tasks involving natural language understanding. In contrast, decoder-only models are designed for text generation, making them ideal for natural language generation tasks. Encoder-decoder models, on the other hand, are adept at sequence-to-sequence conversions, enabling them to transform one form of textual input into another. The inclusion of an encoder in these models allows for contextual understanding, as the context from prior tokens in the input sequence is utilized in generating responses \cite{brown2020language, devlin2018bert, howard2018universal}. These capabilities make LLMs highly versatile and adaptable to a wide range of applications.

BERT has led to various extensions and adaptations. The scientific community has developed numerous BERT variants to address diverse tasks and domains\cite{liu2021robustly}, \cite{lan2019albert}, \cite{sanh2019distilbert}, \cite{rogers2021primer}, \cite{lee2020biobert}.
MechGPT is a specialized variant of the GPT architecture designed for applications in mechanical engineering and related fields \cite{buehler2024mechgpt} \cite{jadhav2024large}. LLM-3D Print is a framework developed to monitor the printing and control processes using pretrained LLMs to determine the parameters causing 3D print failures and autonomously correct them without the need for human intervention \cite{jadhav2024large}. AMGPT represents a significant advancement in the application of AI to additive manufacturing, highlighting the potential of specialized GPT models to transform industry-specific processes and practices \cite{chandrasekhar2024amgpt}.

RAG (Retrieval Augmented Generation)\cite{lewis2020retrieval} is a cutting-edge technique in NLP that enhances LLM performance by combining retrieval-based and generation-based methods. In RAG, a model first retrieves relevant documents from a large corpus using a retriever component, then uses these documents to generate more accurate and contextually relevant responses with a generator component. This hybrid approach significantly improves the model's ability to produce factually accurate and context-rich outputs, addressing limitations in purely generative models by grounding responses in real-world data. RAG is widely adopted in applications requiring high precision and contextual understanding.
RAG offers a groundbreaking approach in tailoring LLMs for specific tasks by retrieving relevant text data for user queries, enhancing the base functionality of LLMs with specialized knowledge. For instance, users can ask questions like "How can we have fast water desalination by maintaining the ion rejection  using 2D nanomaterials?" \cite{barati2024fast} and LLMs can provide intelligent insights with reasonable references comparable to those of human experts.

Nanotechnology and nanoscience are poised to drive the next industrial revolution, fundamentally transforming how we design, manufacture, and innovate. This atomic and molecular approach enables the creation of biologically, chemically, and physically stable structures one atom or molecule at a time, impacting fields such as aerospace, agriculture, defense, energy, environment, materials, manufacturing, and medicine \cite{yousaf2008nanoscience}. The promise of nanoscale science lies in the demonstrated fact that nanoscale materials exhibit mechanical, optical, chemical, and electrical properties that are vastly different from their bulk counterparts \cite{mansoori2017introduction}. For example, macromolecules and particles within the 1–50 nm range possess unique chemical properties such as reactivity and catalytic potential, as well as distinct physical properties such as magnetism and optical behavior. As an enabling technology, nanotechnology improves the durability, reactivity, and performance of materials, allowing the development of products that are smaller, lighter, and stronger. Cutting-edge research areas, including nanolithography, nanodevices, nanorobotics, nanocomputers, nanopowders, and nanostructured catalysts, are advancing rapidly \cite{mansoori2017introduction}. Applications such as molecular manufacturing, nanomedicine for diseases such as Alzheimer's \cite{nazem2008nanotechnology} and cancer\cite{ebrahimi2014reliability}, and innovative drug delivery systems (e.g. liposome vesicles) further demonstrate its potential \cite{de2019drug}. The surge in global investments, research initiatives, and conferences underscores the growing interest in this field, which is poised to revolutionize industries and bring transformative innovations to the way we live and work.

The application of LLMs to nanotechnology can play a pivotal role in parsing the vast amount of scientific literature, helping to discover and design new nanomaterials, facilitating the interpretation of complex experimental data, extracting relevant insights, and predicting the properties of nanomaterials through textual analysis \cite{cao2024machine}. Moreover, the ability of LLMs to generate detailed descriptions and hypotheses offers a powerful tool for conceptualizing new experiments and applications in nanotechnology.

In this work, we explore the application of RAG within the context of nanotechnology, focusing on generating scientific nanotechnology text, accelerating nanomaterial discovery, enhancing literature review processes. We present methods for effectively adapting LLMs to the unique linguistic and conceptual challenges of nanotechnology and discuss the potential chemical and physical considerations and limitations associated with their use. To support further exploration, we provide curated references within each query, guiding readers to relevant scientific paper. Through these efforts, we aim to demonstrate the transformative potential of LLMs in advancing research in the nanotechnology domain.

\section{Related Works}
\label{sec:relatedworks}
\subsection{Ontological Knowledge graphs and Retrieval Augmented Generation}

Ontological Knowledge Graphs (OKGs) have been employed to enhance the interpretability and accuracy of LLMs. By structuring knowledge in a graph format, these systems can provide deeper insights into the relationships between concepts and data points. Buehler et al.\cite{buehlerokg2024} demonstrated the utility of OKGs in materials science, particularly for interpreting and predicting material properties and behaviors. The integration of OKGs with LLMs enables the models to not only retrieve but also contextualize and relate information in a more meaningful way, significantly improving generative performance and reducing hallucinations.

Recent advancements have also explored the use of multi-agent systems, where multiple LLMs with specialized capabilities collaborate to solve complex problems. These systems leverage the strengths of individual agents, such as specific domain expertise or particular functionalities like code generation and execution. Buehler's work highlights the efficacy of multi-agent strategies in materials design and other engineering tasks. By using a combination of agents, each equipped with distinct roles and access to specialized data sources, these systems can perform intricate tasks ranging from hypothesis generation to simulation-based data analysis.

Buehler demonstrated the efficacy of RAG in enhancing the query-answering capabilities of LLMs on specialized topics not covered during their fine-tuning. By integrating contextually relevant information from external knowledge sources, RAG significantly reduces the likelihood of hallucinations and inaccuracies compared to raw LLM responses. This approach allows models to provide more precise and context-rich answers, particularly in fields like nanotechnology, where Buehler showcased improved performance and interpretability using Ontological Knowledge Graphs alongside RAG techniques. 

In the context of nanotechnology, the combination of LLMs and RAG, augmented by OKGs and multi-agent systems, presents a powerful approach to addressing complex research queries. Nanotechnology research often involves interdisciplinary knowledge spanning chemistry, physics, and materials science, necessitating sophisticated tools for information retrieval and integration. 
NANOGPT exemplifies the potential of these advanced AI techniques. Unlike Buehler's work, NANOGPT differentiates itself through the use of a sophisticated real-time retrieval system to provide accurate, contextually enriched responses to intricate queries. While NANOGPT does not currently utilize OKGs, incorporating OKGs into its framework represents an exciting avenue for future research, promising even greater advances in supporting nanotechnology discovery.

\subsection{Large Language Models in Chemistry}

The exploration of LLMs and generative artificial intelligence (GAI) in educational contexts, particularly in the domain of chemistry, has garnered significant attention in recent years. Yik et al.\cite{yik2024chatgpt} contributes to this growing body of literature by investigating the capabilities and limitations of ChatGPT-3.5 in explaining organic chemistry reaction mechanisms.

In chemistry education, the use of GAI tools has been examined primarily through their performance on chemistry assessments. Studies such as those by Leon and Vidhani(2023)\cite{leon2023chatgpt} and Clark(2023)\cite{clark2023investigating} evaluated ChatGPT's ability to answer general chemistry multiple-choice and open-response questions, revealing varied performance with notable challenges in complex problem-solving scenarios. For example, Leon and Vidhani reported that ChatGPT-3.5 achieved a mean score of 27\% on multiple-choice and free response general chemistry questions, while Clark found an accuracy of 47\% on non-numeric questions. These findings underscore the mixed efficacy of LLMs in accurately responding to academic queries.

The application of GAI in organic chemistry, however, has been less explored. Watts et al.\cite{watts2023comparing} conducted a study comparing student-generated and ChatGPT-generated responses to organic chemistry writing-to-learn assignments. Their analysis indicated that, while ChatGPT could engage in mechanistic reasoning, it often lacked the depth and accuracy found in student responses, particularly in describing electron movement, a crucial aspect of mechanistic reasoning. The study builds on this foundation by specifically examining ChatGPT's performance in explaining organic chemistry reaction mechanisms. It evaluates the accuracy and sophistication of the generated explanations, situating the findings within the frameworks of mechanistic reasoning and prompt engineering. The research highlights that while ChatGPT can produce highly sophisticated explanations, accuracy remains a challenge, with a significant portion of responses containing minor inaccuracies that can mislead learners.
NANOGPT, which integrates retrieval-augmented generation to draw directly from verified sources, can significantly reduce hallucinations and enhance the accuracy of generated explanations compared to standalone LLMs like ChatGPT.

\subsection{Natural Language Processing for Materials Science}

Recent advancements in NLP have opened new avenues for analyzing and extracting information from scientific literature in materials science. Choudhary and Kelley\cite{choudhary2023chemnlp} introduced ChemNLP, a comprehensive library for applying NLP techniques to materials chemistry text data. ChemNLP offers several key functionalities relevant to our work. The library provides curated open-access datasets from arXiv and PubChem, specifically tailored for materials and chemistry literature. This approach to creating domain-specific datasets aligns with our goal of developing a specialized system for nanotechnology research.
ChemNLP implements various machine learning models, including traditional algorithms (e.g., SVM, Random Forest), transformers, and graph neural networks, for classifying and clustering scientific texts. The library also incorporates Named Entity Recognition models trained on materials science data, capable of extracting entities such as material names, properties, and characterization methods. This functionality is particularly relevant for identifying key concepts and entities in nanotechnology literature.
The ChemNLP project explores the use of large language models like T5 for abstractive summarization and OPT for text generation in the context of scientific articles. Furthermore, ChemNLP demonstrates the potential of integrating NLP-derived insights with existing materials databases, such as density functional theory (DFT) datasets. This integration showcases the possibility of combining text-based knowledge with structured scientific data, a concept that could enhance the capabilities of our retrieval-augmented generation system.
The development of a web application for chemical formula searches in the ChemNLP project highlights the importance of user-friendly interfaces for accessing processed scientific information, a feature we aim to incorporate in NANOGPT. Overall, the ChemNLP library's approach to processing materials science literature provides valuable insights for our work on NANOGPT, particularly in areas of dataset preparation, entity recognition, and the potential for integrating text-based and structured scientific data in the field of nanotechnology.

\section{Methods}
\label{sec:methods}

\subsection{LLM}

The LLM-RAG system presented in this work utilizes the LLaMA3.1-8b-Instruct model, as described in \cite{dubey2024llama3herdmodels}, due to its sophisticated ability to comprehend and produce natural language. Recognized for its efficiency in balancing computational demands with high performance across a wide range of NLP tasks, LLaMA3.1-8b-Instruct forms the core of our generation process. The model is implemented through the Hugging Face Transformers library, allowing seamless integration of pre-trained models and support for customized workflows. With the transformers library, users gain access to essential resources for working with LLaMA3.1-8B, such as pre-trained model weights and tokenizer setups. Meanwhile, the datasets library plays a key role in managing the retrieval corpus, allowing for streamlined indexing and querying operations that enhance the real-time information retrieval capabilities essential to the RAG model.

\subsection{Retrieval Mechanism}
\begin{figure}[hbt!]

\includegraphics[width=1\linewidth]{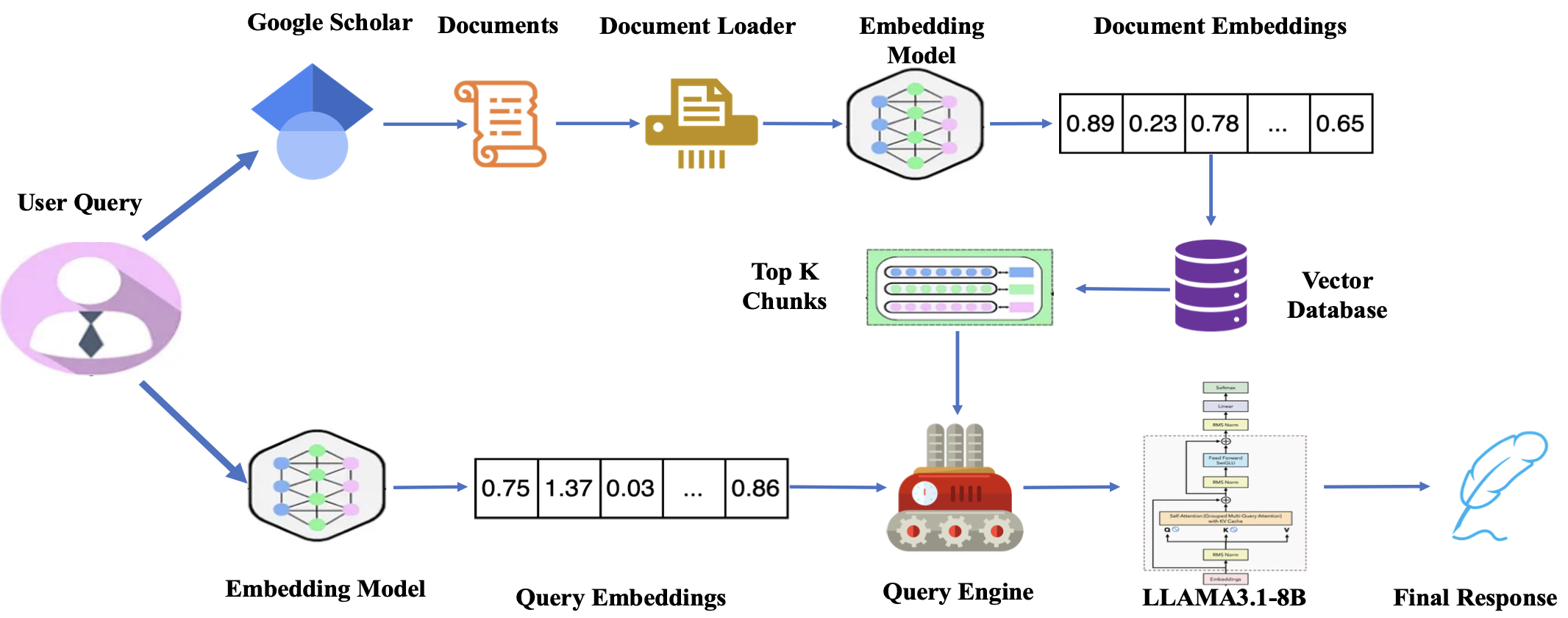}
\caption{RAG Mechanism - the process of querying a database using an embedding model to provide context to an LLM, which in turn generates an answer.
}
\label{fig:transformer_encoders}
\end{figure}

The RAG system presented in this work implements a dual-embedding retrieval mechanism, utilizing a query embedding and a document embedding. Both embeddings are generated from fine-tuned transformer-based models, optimized to represent text inputs as high-dimensional vectors. The query embedding transforms the input prompt into a vector, while the document embedding maps documents from a pre-established corpus into vectors within the same semantic space. This setup allows for the efficient retrieval of the most relevant documents based on the cosine similarity \cite{mikolov2013efficient} between the query and document vectors. This metric involves calculating the cosine of the angle between the query embedding and each indexed embedding. This is done by taking the inner product of the query vector and an indexed vector, normalized by the product of their magnitudes. This normalization focuses on the directionality of the vectors, making it scale-invariant and suited for high-dimensional data comparisons. Cosine similarity scores range from -1 (perfect dissimilarity) to 1 (perfect similarity), with 0 indicating no similarity. The search process retrieves items with the highest positive cosine similarity scores, identifying the most relevant items in the index.

Additionally, our system incorporates a re-ranking process that refines the retrieved documents based on their contextual fit to the query, improving overall retrieval accuracy. The system is also designed to dynamically update the corpus with new information, ensuring it remains flexible and scalable for evolving datasets and diverse use cases.

\subsection{Embedding Models and Semantic Search}

\begin{figure}[hbt!]
\includegraphics[width=1\linewidth]{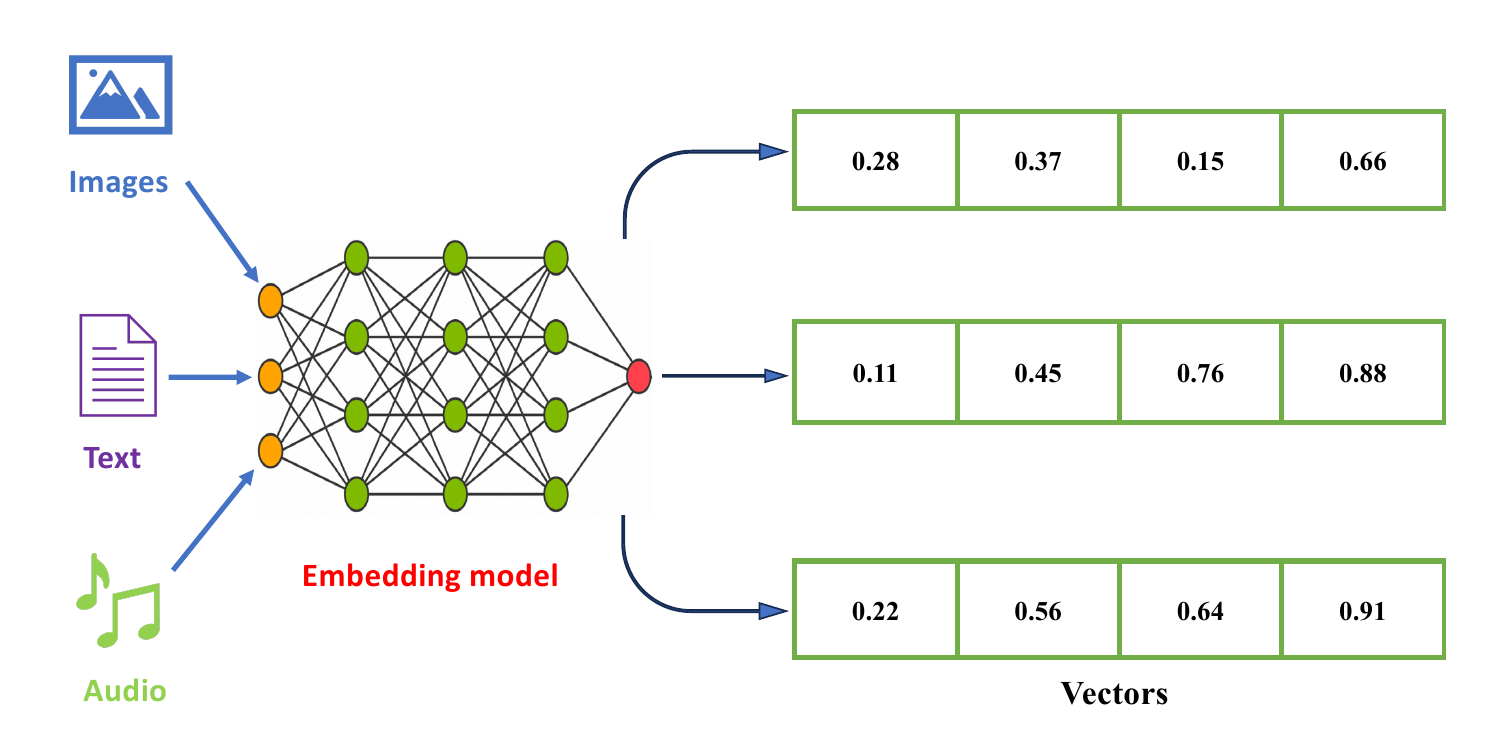}
\caption{
        \textbf{Embedding Model for Multi-Modal Data.} 
        This diagram illustrates how an embedding model transforms various input types: images, text, and audio into numerical vector representations. Each data type is fed into the neural network, which generates unique vectors capturing essential features for further analysis or comparison. The vectors consist of numerical values representing the high-dimensional features encoded from the original input.
    }
\label{fig:emb_model}
\end{figure}

In order to encode natural language into numeric input, an embedding model transforms strings to a high dimensional vector space as shown in Figure \ref{fig:emb_model}. Generally, embeddings can convert any data space into a vectorized representation of each element, enabling multi-modal applications with data types varying from images to audio. For this paper, massive text embeddings are the primary focus.

\subsubsection{Model Selection and Benchmarking}

Semantic search enhances traditional keyword search by focusing on understanding the intent behind a user's query \cite{weisemantic}. This approach relies heavily on embedding models, which form the core of semantic search systems. A key tool for implementing these models is the Python library SentenceTransformers \cite{reimers-gurevych-2019-sentence}, which provides a collection of BERT-based models \cite{devlin-etal-2019-bert} fine-tuned specifically for semantic search tasks.

For this research, we utilize the "sentence-transformers/all-mpnet-base-v2" model, as shown in Figure \ref{fig:emb_model}. This model generates embeddings with a fixed dimensionality of 768, meaning each sentence is encoded as a vector of this length. The model can handle up to 4096 tokens in a single document, and longer documents are split into smaller chunks. These chunks can then be embedded independently or combined, depending on the specific requirements of the task.

In selecting the appropriate embedding model, we referred to the Massive Text Embedding Benchmark (MTEB) by Muennighoff et al. \cite{Muennighoff2022MTEBMT}, which evaluates a wide range of models. The benchmark reveals that no single model excels across all tasks, reflecting the current limitations in the field. Based on our performance requirements and GPU constraints, we selected the "sentence-transformers/all-mpnet-base-v2" model for its balance between efficiency and quality, as validated by the Hugging Face community.

\subsubsection{MPNet Architecture and Contextual Embeddings}

The sentence-transformers/all-mpnet-base-v2 model is built on the MPNet (Masked and Permuted Network) architecture \cite{MPNet}, which advances previous transformer models by combining masked and permuted language modeling techniques. This allows the model to capture dependencies and relationships within the text more effectively, generating richer and more contextually informed embeddings.

Unlike static embeddings, which remain unchanged regardless of context, MPNet produces \textit{contextual embeddings}. These embeddings adapt based on the surrounding text, offering a more accurate representation of word and sentence meaning within a given context.

\subsubsection{Tokenization}

For token counting, we use the cl100k tokenizer from tiktoken \cite{openai_tiktoken_2023}, which is integrated into LlamaIndex. This tokenizer supports a vocabulary of 100,000 tokens and is optimized for compatibility with models such as GPT-4o, making it suitable for processing diverse text inputs with high efficiency.

\subsection{Devices and Codebase}

We employed the LLaMA3.1-8b-Instruct chat model and the  sentence-transformers/all-mpnet-base-v2 embedding model, both operating on two separate local NVIDIA RTX A6000 GPUs, each having 48GB of memory. For public benefit and further research the code is available at the following link: \url{https://github.com/BaratiLab/NANOGPT}.

\vspace{15pt}

%\newpage

\subsection{Literature Extraction}

In this study, we employed a multi-step literature extraction process that integrates various automated retrieval techniques to efficiently gather relevant academic papers.

We utilized the ScienceDirect Search API V2, provided by the Elsevier Developer Portal, to retrieve full-text papers directly from the Elsevier database. This API was leveraged to query papers based on specific identifiers that were extracted using a string search query on Google Scholar. To automate this process, we employed the Selenium open-source python framework\cite{selenium}, which facilitated the scraping of relevant identifiers from Google Scholar search results.

Additionally, Selenium was used to conduct advanced searches on Google Scholar to locate open access papers available on Springer journals. The links to these open access articles were then extracted and subsequently scraped for their full text content. Similarly, we conducted searches on the American Chemical Society (ACS) platform using Selenium to identify open access papers through its advanced search functionalities. The resulting links to accessible papers were also scraped to obtain their textual content.

By integrating these diverse data sources: ScienceDirect, Springer and ACS we were able to compile a rich and diverse dataset of scientific literature. This dataset was then used in subsequent stages of our research, providing a robust foundation for the application of our RAG system.

\newpage

\subsection{User Chat Interface}

\begin{figure}[hbt!]
\centering
\includegraphics[width=0.6\linewidth]{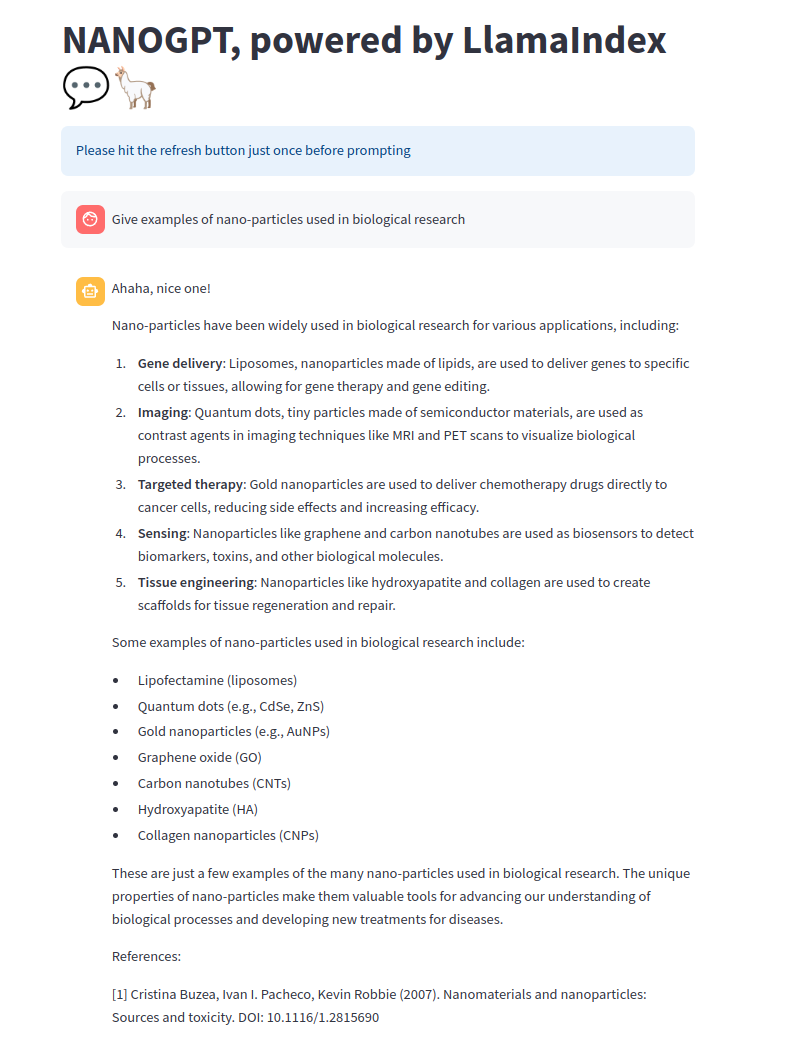}
\caption{Streamlit-based Chat Interface for NANOGPT}
\label{fig:chat_stmlit}
\end{figure}

Streamlit, an open-source Python framework, streamlines the development of web applications, particularly for machine learning and data science projects, by enabling rapid creation of user interfaces without requiring in-depth web development expertise. In this project, it was utilized to design an intuitive chat interface (illustrated in Figure \ref{fig:chat_stmlit}, allowing users to effortlessly interact with a deep learning model that generates responses based on textual or multimodal inputs. Furthermore, Streamlit's functionality includes the ability to store chat history, promoting better continuity in conversations and improving the relevance of future interactions.

\newpage

\subsection{Impact of Maximum Token Length}

The maximum token length parameter sets the limit on the number of tokens that the language model can utilize when generating a response. When the token limit is relatively low (less than 150), the model is constrained to provide brief responses, often no longer than a few sentences. On the other hand, increasing the token limit (beyond 300 tokens) allows the model to produce more detailed and expansive answers, which is beneficial for handling complex or highly descriptive queries. However, simple, fact-based answers that require just a few words or a sentence remain largely unaffected by this parameter. Extremely high token limits (1024 tokens or more) may lead the model to repetitively regurgitate similar content. A maximum token length of 700 was used during the testing of the llama3.1-RAG system.

\subsection{Effect of Sampling Temperature}

Sampling temperature, as described in \cite{hinton}, is a key parameter that controls the level of randomness in the output of a language model during RAG (Retrieval-Augmented Generation) processes. Lower temperatures (below 0.3) prioritize the selection of high-probability tokens, resulting in more conservative and precise outputs. In contrast, higher temperatures (above 1) increase the likelihood of lower-probability token choices, which can lead to more varied and unpredictable results. During tests with the LLaMA3.1-RAG system, a sampling temperature of 0.3 was utilized.

 \subsection{\text{top-k} Retrieval}

In embedding-based search systems, the \textit{top-k} parameter plays a vital role by defining how many of the closest embedding matches are retrieved from the index. It determines the number of relevant results based on similarity scores. Increasing the \textit{top-k} value broadens the search by including more potential matches, which may improve the quality of results.

In the LLaMA3.1-RAG LLM system, setting \textit{top-k} to 3 provided the optimal balance between relevance and conciseness. While a \textit{top-k} value of 2 retrieved fewer examples, higher values like 5 or 6 tended to introduce irrelevant information.

\section{Results} % * syntax makes it not start a new number
% \section*{Results}
\label{sec:results}
\subsection{Evaluation}
The evaluation of NANOGPT was conducted by a panel of experts comprising PhD students and professors who possess significant experience in the application of nanomaterials. This group of evaluators was chosen for their in-depth knowledge of the domain, ensuring that the assessment was both rigorous and relevant to the specialized field of nanotechnology.

To objectively assess the performance of NANOGPT, a ground truth answer was prepared for each query. These ground truth answers were meticulously crafted by human experts to reflect accurate and comprehensive responses to the questions posed, serving as a benchmark for comparison.

The evaluation compared the responses generated by NANOGPT with those generated by the model: Meta-LLaMA/Meta-Llama-3.1-8B-Instruct (without retrieval-augmented generation enabled). Each system was tasked with answering the same set of queries, and their responses were then evaluated against the output of the raw AI model and the ground truth human responses.

The comparison focused on several key metrics, including Depth of Information, Technical Focus, Structure and Clarity, Applications and Specificity, Inclusion of Challenges (where applicable), References, Forward-Looking Insights, and Summary, of the answers provided by each model. These metrics were critical in determining how well each system understood and addressed the complex queries in the context of nanotechnology research.
The results of the evaluation of the RAG system are documented in the appendix.

In 100$\%$ of the responses NANOGPT has outperformed the vanilla non-RAG LLM in technical accuracy and depth. However, the vanilla non-RAG LLM does a better job of explaining scientiific concepts in a simplified manner to a layman.

\newpage

\begin{singlespace}

\label{tab:query-generation}

\end{singlespace}

\newpage

\section{Conclusion and Future Work}
This work introduces NANOGPT, an advanced query-driven, retrieval-augmented generation (RAG) system designed to address the specialized needs of nanotechnology research. By integrating the robust capabilities of the LLaMA3.1-8B-Instruct model with a sophisticated multi-source retrieval mechanism, NANOGPT enhances the efficiency and precision of literature reviews, facilitating faster knowledge acquisition and hypothesis generation. The model's use of embedding-based retrieval, combined with rigorous fine-tuning, demonstrates its ability to provide contextually relevant and accurate insights, streamlining the research workflow for complex nanotechnology queries.

Our evaluations validate NANOGPT’s effectiveness in outperforming traditional models by achieving higher accuracy and relevance in responses, particularly in domain-specific applications. By reducing the cognitive and temporal burden on researchers, NANOGPT showcases the transformative potential of AI-powered tools for the querying of scientific literature. 

Future work will focus on further refining the retrieval mechanisms to integrate dynamic updates from newer datasets, expanding the scope of supported scientific domains, and addressing current limitations, such as managing model hallucinations and optimizing token efficiency. Additionally, incorporating advanced interpretability features and domain-specific ontological knowledge graphs will enhance the transparency and reliability of model outputs. This work represents a significant step toward leveraging AI in nanotechnology research and sets a foundation for broader applications in interdisciplinary scientific domains.

\label{sec:conclusion}

\section*{\textbf{Acknowledgments}}
\label{sec:acknowledgements}

 \bibliographystyle{elsarticle-num} 
 \bibliography{cas-refs}

 \newpage

 \section*{\Huge Appendix: Supplementary Information}
\label{sec:appendix}
\begin{singlespace}

To evaluate the accuracy and precision of word embeddings in the NANOGPT application, we compared its performance on various tasks in the nanotechnology domain against human-generated answers and a well-known generative pretrained model, Groq. We further incorporated author-defined metrics across key subcategories to systematically assess the strengths and weaknesses of each model from multiple perspectives.
Each comparison metric title was chosen to ensure a comprehensive and balanced evaluation of the Groq and NANOGPT model answers, addressing both their strengths and limitations in various aspects:\\

\noindent1. Depth of Information: To assess how thoroughly each model explains the topic. This is crucial for evaluating whether the response provides a high-level overview or dives into detailed insights, catering to different audience needs.\\
2. Technical Focus: To examine the level of technical sophistication in the answer. This highlights whether the response is suitable for general audiences or professionals and researchers requiring advanced knowledge.\\
3. Structure and Clarity:
To evaluate the readability and organization of the information. Clear and structured answers are essential for effective communication, especially for readers with varying levels of expertise.\\
4. Applications and Specificity:
To determine whether the models provide concrete examples and relate their explanations to practical, real-world uses. Specificity enhances the answer’s relevance and applicability.\\
5. Inclusion of Challenges (not always applicable but used in certain cases):
To check if the models address limitations or challenges associated with the topic. This ensures a balanced perspective, especially for nuanced topics like nanotechnology's risks and regulatory considerations.\\
6. References:
To evaluate credibility and the ability to guide readers to additional resources. Including references enhances reliability, especially for research or academic purposes.\\
7. Forward-Looking Insights:
To analyze whether the responses address future directions, trends, or innovations. Forward-looking insights are important for showcasing the evolving nature of the field and inspiring curiosity.\\
8. Summary:
To provide a concise, side-by-side evaluation, helping readers quickly understand the relative strengths and suitability of each model for specific audiences or purposes.
\rule{\linewidth}{0.7mm}
\noindent\rule{\textwidth}{0.4pt} % Full-width thin line
\end{singlespace}

% Use tabular properly
\newpage
\centering
\textbf{Table 1:} \normalsize Comparison of Nanotechnology Applications: Human-Generated Responses vs. \\
Groq Model and NANOGPT Outputs
% Adding space before the table
%\caption{Data ingestion phase of the Retrieval Augmented Generation process}
\centering
\small{
%\begin{tabularx}{\linewidth}{@{}>{\raggedright\arraybackslash}X>{\raggedright\arraybackslash}X>{\raggedright\arraybackslash}X@{}}
\noindent\rule{\textwidth}{0.6pt} % First line
\vspace{1pt}
\raggedright
\textbf{Query 1:} What are the applications of nanomaterials in industries?
\vspace{1pt}

% \hline
% \begin{tabularx}{\linewidth}{@{}>{\raggedright\arraybackslash}p{1\linewidth}X>{\raggedright\arraybackslash}p{1\linewidth}X>{\raggedright\arraybackslash}p{1\linewidth}X@{}} 

% the first param adjusts first column
% the middle one adjusts last column
% the last argument adjusts middle column

\rule{\linewidth}{0.7mm}
% \centering
% \vspace{5pt}

\textbf{Human Generated Answer:} \\
\begin{singlespace}

\justifying{Nanomaterials are key materials in many industrial applications due to their unique characteristics and tunability. A few representative use cases are as follows:\\
Energy Storage: \\
Battery: The core of battery technology lies in the electrochemical reactions occurring in the electrodes (cathode and anode). These electrodes can be made from various types of nanomaterials. For instance, cathode active materials are typically composed of metal oxides, such as LiCoO$_2$ (lithium cobalt oxide), LiFePO$_4$ (lithium iron phosphate, also known as LFP), or LiNiMnCoO$_2$ (lithium nickel manganese cobalt oxide, also known as NMC). On the other hand, the anodes are generally made from carbon-based materials like graphite or silicon, or a combination of both. Additionally, nanostructured membranes, like nanoporous polymers, work as a separator that determines ion transport and helps improve the overall efficiency, safety, and lifespan of the battery.\\
Fuel cell: Fuel cells convert chemical energy into electrical energy through electrochemical reactions, involving hydrogen. Nanomaterials, such as platinum-based nanoparticles, are often used as catalysts in fuel cells to improve reaction efficiency. Nanostructured membranes and electrodes also play a crucial role in optimizing performance and durability.

Chemical Process:\\
Catalyst: Nanomaterials are widely used as catalysts in many kinds of chemical processes due to their large surface area and high reactivity. For instance, platinum nanoparticles are commonly used in automotive catalytic converters to reduce toxic emissions. Additionally, nanocatalysts are used in the petrochemical industry to refine crude oil into fuels and chemicals.\\
Metal-Organic Frameworks (MOFs): MOFs are a class of porous materials constructed from metal ions and organic linkers. These materials have a high surface area and tunable porosity, making them excellent candidates for catalysis, gas storage, and separation processes. The nanoscale control over the pore size and functionality in MOFs allows for highly selective chemical reactions. 

IT Device:\\
Semiconductor: Nanomaterials are essential in the semiconductor industry, where they are used to manufacture transistors, integrated circuits, and other components of electronic devices. Silicon nanowires, for example, are being explored for use in next-generation transistors, enabling smaller and more efficient devices.\\
Display: Nanomaterials such as quantum dots are revolutionizing display technology. Quantum dots are nanoscale semiconductor particles that emit light of specific wavelengths when excited. They are used to produce vibrant colors and enhance image quality while improving energy efficiency. 
	
Environmental and water industries:

Nanomaterials play a significant role in the environmental and water industries, particularly in air and water purification processes. Two-dimensional (2D) nanomaterials, such as graphene, zeolite, and molybdenum disulfide (MoS$_2$), are commonly used in applications like reverse osmosis (RO) and water desalination.}
\end{singlespace}

\rule{\linewidth}{0.7mm}

\vspace{1pt}
\textbf{Groq Model Answer:} \\
\begin{singlespace}
    
\justifying{Nano materials have a wide range of applications in various industries due to their unique physical and chemical properties. Some of the key applications include:
1- Electronics: Nano materials are used in the production of smaller, faster, and more efficient electronic devices such as transistors, diodes, and memory storage devices.\\
2- Solar Energy: Nano materials are used to enhance the efficiency of solar cells by increasing light absorption and reducing energy losses.\\
3- Aerospace: Nano materials are used to make lightweight and strong components for aircraft and spacecraft, such as composite materials and coatings.\\
4- Biomaterials: Nano materials are used in medical implants, wound dressings, and cancer treatment due to their biocompatibility and ability to target specific cells or tissues.\\
5- Cosmetics: Nano materials are used in skincare and haircare products to improve texture, reduce irritation, and enhance penetration of active ingredients.\\
6- Water Filtration: Nano materials are used to create ultra-fine filters for water treatment and purification.\\
7- Biotechnology: Nano materials are used in bioassays, biosensors, and biological imaging.\\
8- Food packaging: Nano materials are used to enhance the food packaging industry with nano-enhanced coatings and wraps.\\
These are just a few examples of the numerous applications of nano materials in various industries. The unique properties of nano materials make them suitable for a wide range of applications.}
\end{singlespace}

\rule{\linewidth}{0.7mm}

\vspace{1pt}    
\textbf{NANOGPT Model Answer:} \\
\begin{singlespace}

\justifying{According to the provided context information, nanomaterials have various applications in different industries. Some of the specific applications mentioned include:\\
Energy storage and conversion: Sodium-ion batteries, lithium-ion batteries, supercapacitors, and photocatalysts for direct solar energy utilization.\\
Sensors: Mechanical sensors, gas sensors, piezoresistive wearable sensors, photoluminescent sensors, and biosensors.\\
Electronics and Photonics: Electrical contacts, conductive fillers, energy harvesting, ferroelectric materials, piezoelectric materials, and optoelectronics.\\
Catalysis: Electrocatalysis, photocatalysis, CO catalyst, CO$_2$ reduction O$_2$ evolution electrocatalyst, and catalyst for NH$_3$ synthesis from N$_2$.\\
Water purification: Membrane separation, toxic heavy metal Cr(vi) adsorption, lead adsorption, phosphate sequestration, and freestanding membranes for the charge- and size-selective rejection of ions and molecules.\\
Biomedical: Antibacterial, bioimaging, biosensing, therapeutics, and nanomedicine.\\
These are just a few examples of the many applications of nanomaterials in various industries. The context information also mentions the synthesis approaches for nanocomposites, including mechanical milling, physical vapor deposition, co-precipitation, inert gas condensation, microemulsion, and hydrothermal methods.

References:

[1] Shaukat Ali Mazari, Esfandyar Ali, Rashid Abro, Fahad Saleem Ahmed Khan, Israr Ahmed, Mushtaq Ahmed, Sabzoi Nizamuddin, Tahir Hussain Siddiqui, Nazia Hossain, Nabisab Mujawar Mubarak, Asif Shah (2021). Nanomaterials: Applications, waste-handling, environmental toxicities, and future challenges – A review. DOI: 10.1016/j.jece.2021.105028}\\
\rule{\linewidth}{0.7mm}
\noindent\rule{\textwidth}{0.4pt} % Full-width thin line

\noindent\textbf{Author Comments:}\\
\noindent\textbf{Depth of Information:}

The NANOGPT model offers more detailed and domain-specific examples, particularly for energy storage, sensors, catalysis, and water purification. It also touches on synthesis methods, which adds technical depth.
The Groq model is broader and less technical, aimed at providing a general overview.\\
\noindent\textbf{Technical Focus:}

NANOGPT focuses on advanced applications (e.g., electrocatalysis, piezoelectric materials) and includes specific chemical processes, appealing to a more technical audience.
Groq avoids technical jargon and is more accessible to general readers.\\
\noindent\textbf{Structure and Clarity:}

The Groq model organizes information into clearly defined categories, making it easier to skim.
NANOGPT includes exhaustive detail but sacrifices readability due to a denser presentation.\\
\noindent\textbf{Inclusion of Challenges:}

NANOGPT briefly acknowledges challenges and waste-handling in the reference section, giving a more nuanced perspective.
Groq omits these aspects entirely.
References:

NANOGPT provides a reference for credibility, whereas Groq does not.\\
\noindent\textbf{Summary:}

Groq Model is better suited for general audiences seeking a broad understanding.
NANOGPT Model is more technical and detailed, targeting researchers or professionals in the field.

\rule{\linewidth}{0.7mm}
\end{singlespace}
\newpage

\centering
\textbf{Table 2:} \normalsize Comparison of the effect of quantum and materials in Nanotechnology: Human-Generated Responses vs.
Groq Model and NANOGPT Outputs

\centering
\small{}
\noindent\rule{\textwidth}{0.6pt} % First line
\vspace{1pt}
\raggedright
\textbf{Query 2:} How do quantum effects influence the behavior of materials at the nanoscale?
\vspace{1pt}

\rule{\linewidth}{0.7mm}
\vspace{1pt}    
\textbf{Human Generated Answer:} \\
\begin{singlespace}
\justifying{At the nanoscale, quantum effects dominate and influence changes in a material's physical, mechanical, optical, and chemical properties, including chemical reactivity, electrical conductivity, magnetic permeability, melting point, fluorescence, etc. For example, color is defined by a material's interaction with light.}

The scale of nanomaterials is typically between 1 and 100 nanometers, where quantum effects dominate physics. Quantum confinement becomes more pronounced for nanoparticles. Unlike bulk materials, where electrons are treated as wave-like free-moving entities, at the nanoscale, the behavior of electrons is influenced by the particles since the spatial extent of the electron wave function is now comparable to the size of the nanoparticles. This confinement results in discrete energy levels, which affect the optical, electrical, and magnetic properties.

\end{singlespace}

\rule{\linewidth}{0.7mm}

\vspace{1pt}
\textbf{Groq Model Answer:} \\
\begin{singlespace}
\justifying{Quantum effects significantly influence the behavior of materials at the nanoscale due to the unique properties that emerge at this size range. Some key effects include:
Quantum confinement: As the size of a material decreases, its energy levels become quantized, leading to changes in its electronic and optical properties.
Tunneling: At the nanoscale, particles can tunnel through potential barriers, allowing for unique transport properties.
Quantum fluctuations: Random fluctuations in energy can occur at the nanoscale, influencing material behavior.
Surface effects: The increased surface-to-volume ratio at the nanoscale can lead to changes in material properties, such as reactivity and conductivity.
Size-dependent properties: Materials can exhibit size-dependent properties, such as changes in melting point, conductivity, or optical properties.
These quantum effects can lead to novel material properties, such as:
- Enhanced conductivity.
- Increased reactivity.
- Improved optical properties.
- Unique magnetic properties.
Understanding these effects is crucial for designing and optimizing nanoscale materials for various applications, including electronics, energy storage, and biomedical devices.}

\end{singlespace}

\rule{\linewidth}{0.7mm}
\vspace{1pt}    
\textbf{NANOGPT Model Answer:} \\
\begin{singlespace}

\justifying{Quantum effects significantly influence the behavior of materials at the nanoscale by altering their physical and chemical properties. As materials decrease in size, they exhibit unique phenomena that differ from their bulk counterparts. For instance, the increased surface area relative to volume leads to heightened reactivity, as a larger percentage of atoms are located on the surface. Additionally, quantum effects such as tunneling allow particles to pass through barriers that would typically repel larger, classical particles. This can enable innovative applications, such as in scanning tunneling microscopy, but can also pose challenges in nano-electronic devices. Overall, these quantum effects create a distinct set of behaviors and properties that are crucial for the development and application of nanomaterials.

References:

[1] Freddy C. Adams, Carlo Barbante (2013). Nanoscience, nanotechnology and spectrometry. DOI: 10.1016/j.sab.2013.04.008}\\
\rule{\linewidth}{0.7mm}
\noindent\rule{\textwidth}{0.4pt} % Full-width thin line

\noindent\textbf{Author Comments:}\\
\noindent\textbf{Depth of Information:}

NANOGPT provides a more nuanced explanation of quantum effects, including phenomena like heightened reactivity due to surface-to-volume ratio and tunneling with specific examples like scanning tunneling applications.
Groq focuses on broader categories (e.g., quantum confinement, surface effects) without specific examples or deeper insights.\\
\noindent\textbf{Technical Focus:}

NANOGPT delves into the scientific implications of quantum effects, such as challenges in nano-electronic devices, which adds depth for technical audiences.
Groq simplifies the discussion, making it more suitable for a general audience.\\
\noindent\textbf{Structure and Clarity:}

The Groq model organizes the content in a clear and categorized manner, aiding readability.
NANOGPT is more verbose and dense, which can make it harder to skim.\\
\noindent\textbf{Inclusion of Applications:}

Both models touch on applications, but NANOGPT integrates the effects more directly into examples, such as scanning tunneling.
Groq connects the effects to general application fields (e.g., electronics, energy storage) without specific cases.\\
\noindent\textbf{Summary:}

Groq Model: Clear and accessible, suitable for a general understanding but lacks depth and references.
NANOGPT Model: Offers detailed insights and examples, better for technical audiences or those seeking a deeper understanding.

\rule{\linewidth}{0.7mm}

\end{singlespace}

\newpage

\centering
\textbf{Table 3:} \normalsize Comparison of Chemical in Nanotechnology: Human-Generated Responses vs. \\
Groq Model and NANOGPT Outputs

\centering
\small{}
\noindent\rule{\textwidth}{0.6pt} % First line
\vspace{1pt}
\raggedright
\textbf{Query 3:} How do surface properties and size affect the chemical reactivity of nanoparticles? What is the suitable surface size of nanomaterials?

\vspace{1pt}

\rule{\linewidth}{0.7mm}
\vspace{1pt}    
\textbf{Human Generated Answer:} \\
\begin{singlespace}

\justifying{Nanoparticle-based materials possess a significantly larger surface area relative to their volume compared to materials composed of larger particles. As a result, the surface plays a crucial role in defining the properties of nanomaterials. Unlike atoms within the bulk, surface atoms are not fully connected to neighboring atoms, making them less stable and higher in energy. Consequently, these surface atoms are more likely to interact with other substances. In essence, nanoparticles, due to their high surface energy and large surface area, tend to be more chemically reactive.

Nanoparticles refer to materials with dimensions in the nanoscale range (1–100 nm). There is a direct correlation between particle size and surface area: as particle size decreases, the specific surface area rises. Additionally, if a particle contains pores, its surface area will further increase due to the additional internal surfaces exposed (Figure \ref{fig:q3}S)}.
\begin{figure}[hbt!]
\includegraphics[width=1\linewidth]{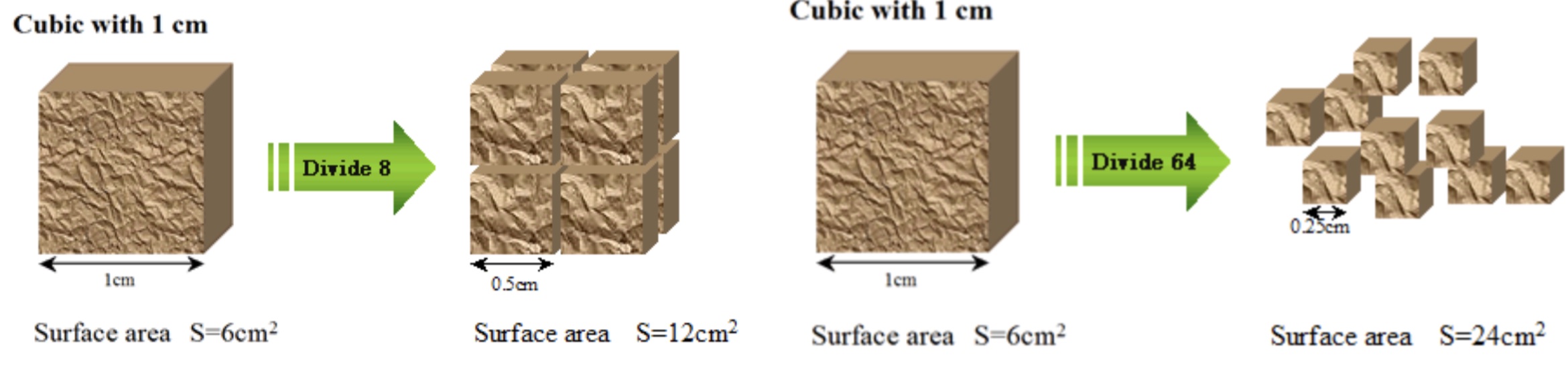}
\caption{Effects of slicing the surfaces on surface area}
\label{fig:q3}
\end{figure}
\end{singlespace}
\rule{\linewidth}{0.7mm}
\vspace{0.3pt}
\textbf{Groq Model Answer:} \\
\begin{singlespace}

\justifying{Surface properties and size significantly affect the chemical reactivity of nanoparticles due to their unique physical and chemical characteristics. Key factors include:
Surface area-to-volume ratio: As particle size decreases, the surface area increases, making nanoparticles more reactive.
Surface energy: High surface energy of nanoparticles can lead to increased reactivity, as atoms on the surface are more loosely bound and more prone to chemical reactions.
Defects and impurities: Nanoparticles often have defects and impurities on their surface, which can increase their reactivity.
The suitable surface size of nanomaterials depends on the application:
Biomedical applications: Nanoparticles with diameters between 10-100 nm are often used, as they can be easily taken up by cells.
Energy applications: Nanoparticles with diameters between 1-10 nm are often used, as they can exhibit enhanced catalytic properties.
Optical applications: Nanoparticles with diameters between 10-100 nm are often used, as they can exhibit unique optical properties.
In general, nanoparticles with diameters between 1-100 nm are often used in various applications, as they exhibit unique physical and chemical properties that are not seen in larger particles.}

\end{singlespace}

\rule{\linewidth}{0.7mm}
\vspace{1pt}    
\textbf{NANOGPT Model Answer:} \\
\begin{singlespace}

\justifying{Surface properties and size of nanoparticles can significantly affect their chemical reactivity. The surface properties of nanoparticles, such as their charge, hydrophobicity, and surface roughness, can influence their interactions with other molecules and their reactivity. For example, nanoparticles with a hydrophilic surface may be more reactive than those with a hydrophobic surface.

The size of nanoparticles can also affect their chemical reactivity. As the size of nanoparticles decreases, their surface area-to-volume ratio increases, which can lead to increased reactivity. This is because smaller nanoparticles have a larger surface area that is exposed to the surrounding environment, making them more susceptible to chemical reactions.

Regarding the suitable surface size of nano materials, it depends on the specific application and the desired properties. In general, nanoparticles with a size range of 1-10 nm are considered to be in the "sweet spot" for many applications, as they have a large surface area-to-volume ratio and are still small enough to be easily dispersed in a variety of solvents.

However, it's worth noting that the surface size of nanoparticles can also affect their stability and aggregation behavior. For example, nanoparticles with a larger surface area may be more prone to aggregation and less stable than those with a smaller surface area.
In summary, the surface properties and size of the nanoparticles can significantly affect their chemical reactivity, and the appropriate surface size of the nanomaterials depends on the specific application and desired properties.

References:

[1] Ferronato et al. (2023) Environmental Life Cycle Assessment of biomass and cardboard
waste-based briquettes production and consumption in Andean areas. 10.1016/j.apenergy.2017.01.07

[2] Nouha Bakaraki Turan, Hanife Sari Erkan, Guleda Onkal Engin, Mehmet Sinan Bilgili Nanoparticles in the aquatic environment: Usage, properties, transformation and toxicity—A review (2019) 10.1016/j.psep.2019.08.014
}\\
\rule{\linewidth}{0.7mm}
\noindent\rule{\textwidth}{0.4pt} % Full-width thin line

\noindent\textbf{Author Comments:}\\
\noindent\textbf{Depth of Information:}

NANOGPT delves deeper into surface properties, addressing factors like charge, hydrophobicity, and surface roughness, which affect reactivity and interactions.
Groq focuses on broader aspects such as surface energy and defects but omits finer details about surface interactions.\\
\noindent\textbf{Technical Focus:}

NANOGPT highlights challenges such as stability and aggregation due to high surface area, adding practical considerations.
Groq emphasizes the versatility of nanoparticles across applications but does not address potential drawbacks.\\
\noindent\textbf{Application Specificity:}

Both models link nanoparticle size to specific applications, but NANOGPT provides a more detailed rationale, such as the "sweet spot" size range of 1-10 nm for optimal reactivity and dispersibility.
Groq provides size ranges but in a more generalized way, focusing on broader categories.\\
\noindent\textbf{Structure and Clarity:}

Groq is highly structured, with clear categorization of applications based on nanoparticle size.
NANOGPT is less structured but more descriptive, offering richer explanations at the cost of readability.\\
\noindent\textbf{References:}

NANOGPT includes references to support its claims and offer credibility.
Groq lacks citations, which limits its authority on the topic.\\
\noindent\textbf{Summary:}

Groq Model: Clear, structured, and application-focused, ideal for general audiences seeking an overview.
NANOGPT Model: More detailed and technically insightful, addressing specific surface properties and challenges, making it better suited for technical readers or researchers.

\rule{\linewidth}{0.7mm}
\end{singlespace}

\newpage

\centering
\textbf{Table 4:} \normalsize Comparison of Environment in Nanotechnology: Human-Generated Responses vs. \\
Groq Model and NANOGPT Outputs

\centering
\small{}
\noindent\rule{\textwidth}{0.6pt} % First line
\vspace{1pt}
\raggedright
\textbf{Query 4:} What are the potential environmental and health risks associated with the widespread use of nanomaterials?

\vspace{1pt}

\rule{\linewidth}{0.7mm}
\vspace{1pt}    
\textbf{Human Generated Answer:} \\
\begin{singlespace}
\justifying{Nanomaterials present both remarkable opportunities and significant risks, primarily due to their unique properties. While these properties offer major advantages, they also bring challenges that demand attention. The primary concern is that extensive research is still needed to fully comprehend the environmental and health implications of nanomaterials. Unlike bulk materials, which are relatively well-studied, nanomaterials exhibit distinct characteristics stemming from their large surface area and high surface energy. This makes predicting their long-term effects difficult. Specifically, there is limited understanding of how nanomaterials interact with organisms and chemicals in various environments, as well as how they accumulate within biological systems.

For example, once nanomaterials enter the bloodstream, they can circulate throughout the body, potentially lodging in organs such as the brain, liver, heart, kidneys, spleen, bone marrow, and nervous system. Once inside cells, they may disrupt normal cellular processes, cause oxidative stress, or even lead to cell death.

Moreover, the high reactivity of nanoparticles can result in unexpected reactions, posing risks of fire or explosion. Therefore, meticulous care must be taken during the processing and storage of nanomaterials to mitigate such hazards.}

\end{singlespace}

\rule{\linewidth}{0.7mm}

\vspace{1pt}
\textbf{Groq Model Answer:} \\
\begin{singlespace}
\justifying{The use of nanomaterials has raised concerns about potential environmental and health risks. Some of these risks include:
Toxicity: Nanoparticles can be toxic to humans and the environment due to their small size and high surface area.
Persistence: Nanoparticles can persist in the environment for long periods, causing long-term damage.
Bioaccumulation: Nanoparticles can accumulate in organisms and potentially cause harm.
Respiratory problems: Inhaling nanoparticles can cause respiratory problems, such as lung damage and inflammation.
Skin irritation: Nanoparticles can cause skin irritation and allergic reactions.
Environmental contamination: Nanoparticles can contaminate waterways and soil, affecting ecosystems.
To mitigate these risks, researchers and manufacturers are working to develop safer nanomaterials and better understand their behavior in the environment.}

\end{singlespace}

\rule{\linewidth}{0.7mm}
\vspace{1pt}    
\textbf{NANOGPT Model Answer:} \\
\begin{singlespace}
\justifying{The widespread use of nanomaterials raises several potential environmental and health risks. These include:

Exposure Pathways: Nanomaterials can enter the environment through various pathways, including air, water, and soil, leading to potential exposure for humans and wildlife.

Toxicity: There are concerns regarding the toxicity of nanoparticles, which may have different biological effects compared to their larger counterparts. This includes potential harm to lung cells and other biological systems.

Bioaccumulation: Nanoparticles may accumulate in organisms, leading to biomagnification through food chains, which can impact ecological health and human safety.

Environmental Behavior: Once released, nanoparticles can behave unpredictably, undergoing processes such as agglomeration, degradation, and adsorption, which complicates risk assessment.

Insufficient Data: There is a lack of comprehensive data on the real-time exposure levels of engineered nanomaterials in various environmental matrices, making it difficult to predict their long-term impacts.

Regulatory Challenges: Current risk management techniques may not adequately address the unique properties of nanoparticles, as they often rely on mass rather than surface area or quantity for risk evaluation.

These factors highlight the need for further research and a coherent approach to understanding and managing the risks associated with nanomaterials.

References:

[1] Agnieszka Gajewicz, Bakhtiyor Rasulev, Tandabany C. Dinadayalane, Piotr Urbaszek, Tomasz Puzyn, Danuta Leszczynska, Jerzy Leszczynski (2012). Advancing risk assessment of engineered nanomaterials: Application of computational approaches. DOI: 10.1016/j.addr.2012.05.014

[2] C. Martin, A. Nourian, M. Babaie, G.G. Nasr (2023). Environmental, health and safety assessment of nanoparticle application in drilling mud – Review. DOI: 10.1016/j.geoen.2023.211767}\\
\rule{\linewidth}{0.7mm}
\noindent\rule{\textwidth}{0.4pt} % Full-width thin line

\noindent\textbf{Author Comments:}\\
\noindent\textbf{Depth of Information:}

NANOGPT provides a more detailed discussion, addressing not just the risks but also the mechanisms and challenges, such as unpredictable environmental behavior and insufficient data for risk assessment.
Groq focuses on listing the risks but does not delve into their complexities or implications.\\
\noindent\textbf{Technical Focus:}

NANOGPT includes advanced topics like biomagnification through food chains and regulatory challenges, adding depth for an expert audience.
Groq simplifies the discussion, suitable for a general overview but less informative for technical readers.\\
\noindent\textbf{Structure and Clarity:}

Groq is more structured and easier to read due to its clear categorization of risks.
NANOGPT provides a denser and more detailed narrative, which can be less accessible to non-technical audiences.\\
\noindent\textbf{Inclusion of Solutions:}

Both models acknowledge the need for mitigating risks, but NANOGPT emphasizes the need for further research and regulatory improvements, adding a forward-looking perspective.
Groq briefly mentions efforts to develop safer nanomaterials but does not elaborate.\\
\noindent\textbf{References:}

NANOGPT supports its claims with credible references, enhancing its reliability.
Groq lacks citations, which reduces its scientific credibility.\\
\noindent\textbf{Summary:}

Groq Model: Clear and structured, providing a basic overview of risks, suitable for general audiences.
NANOGPT Model: More detailed and technically insightful, addressing mechanisms, data gaps, and regulatory challenges, ideal for researchers or experts.

\rule{\linewidth}{0.7mm}
\end{singlespace}
\newpage

\centering
\textbf{Table 5:} \normalsize Comparison of Physics query in Nanotechnology: Human-Generated Responses vs. \\
Groq Model and NANOGPT Outputs

\centering
\small{}
\noindent\rule{\textwidth}{0.6pt} % First line
\vspace{1pt}
\raggedright
\textbf{Query 5:} What is the thermal conductivity of nanomaterials?
\vspace{1pt}
\rule{\linewidth}{0.7mm}
\vspace{0.5pt}    
\textbf{Human Generated Answer:} \\
\begin{singlespace}
\justifying{Nanomaterials typically have a very large surface area for heat transfer. Specifically, the large surface exposes plenty of electrons available for heat transfer. Thus, the size, shape, and dimensionality significantly impact the thermal conductivity of nanomaterials. In general, thermal conductivity decreases as the size of nanomaterials decreases. The variation in thermal conductivity with size is calculated for nanowire, nanofilm, spherical, regular tetrahedral, regular octahedral, and regular icosahedral particle shapes.
\url{https://www.sciencedirect.com/science/article/pii/S2214785321033290} Additionally, nanofluids, are known to have excellent heat transfer properties, making them effective coolants. Current physics cannot fully explain the exact reason behind this phenomenon, but the layering of fluid around the nanoparticles is believed to enhance heat transfer by providing additional pathways.}
\end{singlespace}

\rule{\linewidth}{0.7mm}

\vspace{0.5pt}
\textbf{Groq Model Answer:} \\
\begin{singlespace}
\justifying{The thermal conductivity of nanomaterials can vary greatly depending on their composition, structure, and dimensions. Generally, nanomaterials have unique thermal properties compared to their bulk counterparts.
Some common thermal conductivity values for nanomaterials are:
Carbon nanotubes: 2000-6000 W/mK (highly dependent on orientation and structure)
Graphene: 5000 W/mK
Silicon nanowires: 10-100 W/mK
Metal nanoparticles: 100-500 W/mK
Polymer nanofibers: 0.1-10 W/mK
Nanomaterials can exhibit enhanced or reduced thermal conductivity due to their:
High surface area-to-volume ratio
Quantum confinement effects
Phonon scattering at the nanoscale
Structural defects or impurities
These properties make nanomaterials attractive for applications in thermal management, energy storage, and electronic devices. However, the thermal conductivity of nanomaterials can be challenging to predict and measure due to their complex structure and behavior.}

\end{singlespace}

\rule{\linewidth}{0.7mm}
\vspace{0.5pt}    
\textbf{NANOGPT Model Answer:} \\
\begin{singlespace}

\justifying{The thermal conductivity of nano materials can vary greatly depending on the specific material and its properties. For example, the thermal conductivity of TiO2 nanoparticles is around 8.4 W/mK, while the thermal conductivity of MWCNTs (multi-walled carbon nanotubes) can be as high as 2000-3000 W/mK. Other nano materials such as Al$_2$O$_3$, CuO, Fe$_2$O$_3$, MgO, SiO$_2$, and ZnO have thermal conductivities ranging from 1.34-54.9 W/mK. However, it's important to note that the thermal conductivity of nano materials can also depend on the particle size, shape, and concentration, as well as the properties of the base fluid they are dispersed in.

References:

[1] Mohamed M. Tawfik (2017). Experimental studies of nanofluid thermal conductivity enhancement and applications: A review. DOI: 10.1016/j.rser.2016.11.111

[2] M.I. Pryazhnikov, A.V. Minakov, V.Ya. Rudyak, D.V. Guzei (2017). Thermal conductivity measurements of nanofluids. DOI: 10.1016/j.ijheatmasstransfer.2016.09.080
}\\
\rule{\linewidth}{0.7mm}
\noindent\rule{\textwidth}{0.4pt} % Full-width thin line

\noindent\textbf{Author Comments:}\\
\noindent\textbf{Depth of Information:}

NANOGPT offers a more detailed discussion with specific thermal conductivity values for a wider range of nanomaterials (e.g., TiO$_2$, MWCNTs, Al$_2$O$_3$, ZnO) and highlights the influence of factors like particle size, shape, and dispersion medium.
Groq focuses on broad categories (e.g., carbon nanotubes, graphene) with less emphasis on variability or contextual factors.\\
\noindent\textbf{Technical Focus:}

NANOGPT provides a nuanced explanation of the dependence of thermal conductivity on external factors such as base fluid properties and nanoparticle concentration.
Groq highlights general influencing factors like phonon scattering and structural defects but does not dive into their practical implications.\\
\noindent\textbf{Structure and Clarity:}

Groq is well-structured, presenting clear categories and thermal conductivity ranges for common nanomaterials.
NANOGPT is more detailed but less structured, potentially overwhelming for readers without a technical background.\\
\noindent\textbf{Applications:}

Both models briefly mention applications, but Groq connects thermal properties directly to practical uses like energy storage and thermal management, making it more application-focused.
NANOGPT emphasizes the variability and measurement challenges, offering a broader scientific perspective.\\
\noindent\textbf{References:}

NANOGPT includes citations for thermal conductivity studies, adding credibility and avenues for further exploration.
Groq does not provide references, reducing its reliability for academic or research purposes.\\
\noindent\textbf{Summary:}

Groq Model: Clear and structured, ideal for a general audience seeking an overview of thermal conductivity and its relevance to applications.
NANOGPT Model: More detailed and technical, offering specific data points, variability factors, and references, suited for researchers or experts in the field.

\rule{\linewidth}{0.7mm}
\end{singlespace}

\newpage

\centering
\textbf{Table 6:} \normalsize Comparison of drug discovery in Nanotechnology: Human-Generated Responses vs. Groq Model and NANOGPT Outputs
\centering
\small{}
\noindent\rule{\textwidth}{0.6pt} % First line
\vspace{1pt}
\raggedright
\textbf{Query 6:} How can nanotechnology be used to enhance drug delivery systems for targeted therapy?

\vspace{1pt}

\rule{\linewidth}{0.7mm}
\vspace{1pt}    
\textbf{Human Generated Answer:} \\
\begin{singlespace}
\justifying{The small size and large surface area of nanomaterials enhance drug solubility, thereby increasing bioavailability. For example, nano-sized drug particles can cross the blood-brain barrier (BBB), which typically prevents conventional drug molecules from passing through. Similarly, nanoparticles can deliver drugs directly to cancer cells, bypassing mechanisms of drug resistance and increasing treatment effectiveness. In addition to targeted drug delivery, nanoparticles can also be used for imaging, enabling real-time monitoring of treatment progress and disease state.}

\end{singlespace}

\rule{\linewidth}{0.7mm}

\vspace{1pt}
\textbf{Groq Model Answer:} \\
\begin{singlespace}
\justifying{Nanotechnology has the potential to revolutionize drug delivery systems by providing targeted therapy and improving treatment outcomes. Here are some ways nanotechnology can be used:
Targeted delivery: Nanoparticles can be designed to target specific cells or tissues, reducing side effects and improving efficacy.
Controlled release: Nanoparticles can be engineered to release drugs at a controlled rate, providing a sustained therapeutic effect.
Improved bioavailability: Nanoparticles can enhance the solubility and bioavailability of poorly soluble drugs.
Imaging and diagnostics: Nanoparticles can be used as contrast agents for imaging and diagnostics, allowing for real-time monitoring of disease progression and treatment response.
Examples of nanotechnology-based drug delivery systems include:
Liposomes: tiny vesicles that encapsulate drugs and target specific cells.
Nanoparticles: tiny particles made of materials such as gold, silver, or polymers that can be designed to target specific cells or tissues.
Micelles: tiny spheres that encapsulate drugs and target specific cells.
These technologies have the potential to improve treatment outcomes and reduce side effects, making them promising tools for targeted therapy.}

\end{singlespace}

\rule{\linewidth}{0.7mm}
\vspace{1pt}    
\textbf{NANOGPT Model Answer:} \\
\begin{singlespace}
\justifying{Nanotechnology has the potential to revolutionize drug delivery systems for targeted therapy by providing a platform for the design and development of novel, targeted, and controlled release systems. Here are some ways nanotechnology can be used to enhance drug delivery systems:

Targeted delivery: Nanoparticles can be engineered to target specific cells, tissues, or organs, allowing for more efficient and effective delivery of therapeutic agents.
Improved bioavailability: Nanoparticles can enhance the bioavailability of therapeutic agents by increasing their solubility, stability, and absorption rates.
Controlled release: Nanoparticles can be designed to release therapeutic agents in a controlled and sustained manner, reducing the risk of toxicity and improving treatment outcomes.
Enhanced tumor targeting: Nanoparticles can be engineered to target tumors specifically, allowing for more effective treatment of cancer and other diseases.
Improved imaging: Nanoparticles can be used as contrast agents for imaging, allowing for real-time monitoring of drug delivery and treatment outcomes.
Some examples of nanotechnology-based drug delivery systems include:
Liposomes: Liposomes are nanoparticles composed of lipids that can encapsulate therapeutic agents and target specific cells or tissues.
Nanoparticle-based drug delivery systems: These systems use nanoparticles as the carrier for therapeutic agents, allowing for targeted delivery and controlled release.
Nanocarriers: Nanocarriers are nanoparticles that can carry therapeutic agents and target specific cells or tissues.
Nanoparticle-based vaccines: These systems use nanoparticles as the carrier for antigens, allowing for targeted delivery and improved immune responses.
Overall, nanotechnology has the potential to enhance drug delivery systems for targeted therapy by providing a platform for the design and development of novel, targeted, and controlled release systems. By leveraging the unique properties of nanoparticles, researchers can develop more effective and efficient drug delivery systems that improve treatment outcomes and reduce side effects.10.1016/j.jconrel.2011.06.001 10.1016/j.jconrel.2012.07.010

References:

[1] You Han Bae , Kinam Park (2011). Targeted drug delivery to tumors: Myths, reality and possibility DOI: 10.1016/j.jconrel.2011.06.001

[2] Il Keun Kwon ,Sang Cheon Lee, Bumsoo Han, Kinam Park (2012)
Analysis on the current status of targeted drug delivery to tumors DOI: 10.1016/j.jconrel.2012.07.010
}\\
\rule{\linewidth}{0.7mm}
\noindent\rule{\textwidth}{0.4pt} % Full-width thin line

\noindent\textbf{Author Comments:}\\
\noindent\textbf{Depth of Information:}

NANOGPT offers a more detailed explanation, covering enhanced tumor targeting, nanoparticle-based vaccines, and real-time imaging for monitoring drug delivery and outcomes.
Groq focuses on foundational aspects like targeted delivery and controlled release without delving into specific advancements or niche applications.\\
\noindent\textbf{Technical Focus:}

NANOGPT provides a richer discussion of the mechanisms behind improved bioavailability, stability, and absorption, making it more suitable for technical or research-oriented audiences.
Groq simplifies the concepts, making them accessible to general readers but less useful for advanced study.\\
\noindent\textbf{Structure and Clarity:}

Groq is well-structured with bullet points and examples, offering a clear and straightforward presentation.
NANOGPT is more verbose, with denser information that might be harder to digest for non-expert audiences.\\
\noindent\textbf{Applications:}

Both models highlight similar technologies (e.g., liposomes, nanoparticles), but NANOGPT introduces more variety, such as nanoparticle-based vaccines, and adds specificity to applications like tumor targeting.\\
\noindent\textbf{References:}

NANOGPT includes references to credible sources, enhancing its reliability and providing pathways for further exploration.
Groq lacks citations, which weakens its scientific credibility.\\
\noindent\textbf{Forward-Looking Insights:}

NANOGPT emphasizes leveraging nanoparticle properties for novel designs and improved efficiency, projecting a broader potential impact on future drug delivery systems.
Groq focuses more on the current state of technology without addressing future directions.\\
\noindent\textbf{Summary:}

Groq Model: Clear and structured, providing a solid introduction to the role of nanotechnology in drug delivery, suitable for general audiences.
NANOGPT Model: More detailed and technically insightful, with broader applications and references, making it ideal for researchers or professionals in the field.

\rule{\linewidth}{0.7mm}
\end{singlespace}

\centering
\textbf{Table 7:} \normalsize Comparison of environment in Nanotechnology: Human-Generated Responses vs. \\
Groq Model and NANOGPT Outputs
\centering
\small{}
\noindent\rule{\textwidth}{0.6pt} % First line
\vspace{1pt}
\raggedright
\textbf{Query 7:} How can nanotechnology be leveraged to clean up environmental pollutants and toxins?

\vspace{1pt}

\rule{\linewidth}{0.7mm}
\vspace{1pt}    
\textbf{Human Generated Answer:} \\
\begin{singlespace}
\justifying{Nanotechnology introduces advanced environmental solutions, including nanosensors, nanodevices, nanofilters, and nanorobots, which have significant potential for addressing environmental pollution and contamination problems. Nanotechnology can be applied to two high-level categories: pollutant detection and purification. Specifically, nanosensors can detect and monitor contaminants. Nanotechnology enhances sensing capabilities by utilizing nanoscale materials that have high sensitivity due to their large surface area and quantum properties, which allow them to detect even trace amounts of pollutants. These nanosensors can be designed to target specific chemicals, biological agents, or heavy metals, providing real-time data for more accurate monitoring of contamination levels.

Additionally, nanofiltration devices can be used to purify contaminants. Nanomaterials exhibit enhanced reactivity and greater effectiveness compared to their bulk counterparts, primarily due to their higher surface-to-volume ratio. This allows for more efficient interaction with pollutants, facilitating faster and more thorough purification. Furthermore, nanomaterials can be functionalized or grafted with specific chemical groups that target and bind to pollutants, such as heavy metals or organic toxins, enabling selective removal from the environment. These properties make nanotechnology-based filtration systems a powerful tool in environmental remediation efforts, providing more precise and efficient pollutant removal than traditional methods.
}
\end{singlespace}

\rule{\linewidth}{0.7mm}

\vspace{1pt}
\textbf{Groq Model Answer:} \\
\begin{singlespace}
\justifying{Nanotechnology can play a significant role in cleaning up environmental pollutants and toxins. Here are some ways:
Remediation of contaminated soil: Nanoparticles can be designed to target and break down pollutants such as heavy metals, pesticides, and industrial chemicals, making them more easily removable.
Water purification: Nanomaterials can be used to create efficient filters that remove pollutants and contaminants from water, making it safe for human consumption.
Bioremediation: Nanoparticles can be used to enhance the activity of microorganisms that break down pollutants, accelerating the cleanup process.
Advanced sensors: Nanotechnology can be used to develop sensitive sensors that detect pollutants and toxins, allowing for early detection and response.
Targeted delivery: Nanoparticles can be designed to target specific pollutants, allowing for more efficient and effective cleanup.
Overall, nanotechnology has the potential to revolutionize the way we clean up environmental pollutants and toxins, making our environment safer and healthier.
}
\end{singlespace}

\rule{\linewidth}{0.7mm}
\vspace{1pt}    
\textbf{NANOGPT Model Answer:} \\
\begin{singlespace}
\justifying{Nanotechnology can be leveraged to clean up environmental pollutants and toxins through the use of manufactured nanomaterials (MNMs) that are designed to immobilize, transform, and degrade chemical pollutants. These nanomaterials possess unique properties at the nanoscale, such as enhanced catalytic, adsorptive, and antimicrobial capabilities, which make them effective for detecting and removing contaminants from water, soil, and air. Additionally, their ability to migrate easily allows for efficient treatment of large and hard-to-reach areas, improving cost-effectiveness in remediation efforts. However, it is essential to address the safety concerns associated with MNMs to ensure their successful application in environmental cleanup.

References:

[1] Xinghui Liu, Kuppusamy Sathishkumar, Huichao Zhang, Kuldeep K Saxena, Fuchun Zhang, Saraschandra Naraginti, Anbarasu K, Ramya Rajendiran, Aruliah Rajasekar, Xiang Guo (2024). Frontiers in environmental cleanup: Recent advances in remediation of emerging pollutants from soil and water. DOI: 10.1016/j.hazadv.2024.100461

[2] Ilaria Corsi, Iole Venditti, Francesco Trotta, Carlo Punta (2023). Environmental safety of nanotechnologies: The eco-design of manufactured nanomaterials for environmental remediation. DOI: 10.1016/j.scitotenv.2022.161181

\textbf{follow up question: Give examples of these materials}\\

Manufactured nanomaterials (MNMs) used for environmental remediation include a variety of substances designed for specific pollutant interactions. Examples of these materials are:

Nanoparticles of Zero-Valent Iron (nZVI) - These are used for the reduction of contaminants like chlorinated solvents and heavy metals in groundwater.

Titanium Dioxide Nanoparticles (TiO2) - Known for their photocatalytic properties, they are effective in degrading organic pollutants under UV light.

Silver Nanoparticles (AgNPs) - These possess antimicrobial properties and can be used to purify water by eliminating bacteria and other pathogens.

Carbon Nanotubes (CNTs) - Their high surface area and adsorption capabilities make them suitable for capturing organic pollutants from water and air.

Magnetic Nanoparticles - These can be easily separated from treated media using magnetic fields, making them useful for the removal of heavy metals and organic contaminants.

These materials leverage their unique nanoscale properties to enhance the efficiency and effectiveness of pollution detection and removal processes.

References:

[1] Ilaria Corsi, Iole Venditti, Francesco Trotta, Carlo Punta (2023). Environmental safety of nanotechnologies: The eco-design of manufactured nanomaterials for environmental remediation. DOI: 10.1016/j.scitotenv.2022.161181

[2] Xinghui Liu, Kuppusamy Sathishkumar, Huichao Zhang, Kuldeep K Saxena, Fuchun Zhang, Saraschandra Naraginti, Anbarasu K, Ramya Rajendiran, Aruliah Rajasekar, Xiang Guo (2024). Frontiers in environmental cleanup: Recent advances in remediation of emerging pollutants from soil and water. DOI: 10.1016/j.hazadv.2024.100461
}\\
\rule{\linewidth}{0.7mm}
\noindent\rule{\textwidth}{0.4pt} % Full-width thin line

\noindent\textbf{Author Comments:}\\
\noindent\textbf{Depth of Information:}

NANOGPT offers detailed insights into how manufactured nanomaterials (MNMs) are used for environmental cleanup, including specific examples like zero-valent iron nanoparticles (nZVI) for groundwater decontamination and titanium dioxide nanoparticles (TiO$_2$) for photocatalysis.
Groq provides a high-level explanation without naming specific nanomaterials or their mechanisms of action.\\
\noindent\textbf{Technical Focus:}

NANOGPT explains the unique properties of MNMs, such as catalytic, adsorptive, and antimicrobial capabilities, and addresses their mobility and cost-effectiveness in large-scale remediation.
Groq emphasizes broad benefits like targeting specific pollutants and enhancing bioremediation but omits the technical nuances of nanomaterials’ functionality.\\
\noindent\textbf{Structure and Clarity:}

Groq is well-structured and clear, presenting applications in categorized points, making it more accessible to general readers.
NANOGPT is more detailed but denser, which may require a technical background for full comprehension.\\
\noindent\textbf{Applications:}

NANOGPT includes concrete examples of nanomaterials, such as carbon nanotubes for capturing organic pollutants and magnetic nanoparticles for heavy metal removal.
Groq discusses applications at a conceptual level without linking them to specific nanomaterials or examples.\\
\noindent\textbf{References:}

NANOGPT provides references, increasing its credibility and offering avenues for further study.
Groq does not include citations, reducing its utility for research purposes.\\
\noindent\textbf{Forward-Looking Insights:}

NANOGPT briefly mentions the need to address safety concerns and eco-design for successful application, reflecting a more comprehensive perspective on environmental remediation.
Groq focuses on the benefits without addressing potential risks or challenges.\\
\noindent\textbf{Summary:}

Groq Model: Clear and accessible, providing a general introduction to nanotechnology’s role in environmental cleanup, suitable for non-technical audiences.
NANOGPT Model: More detailed and technical, with specific examples, references, and considerations for safety and application design, ideal for researchers or professionals.

\rule{\linewidth}{0.7mm}
\end{singlespace}

\newpage

\centering
\textbf{Table 8:} \normalsize Comparison of products in Nanotechnology: Human-Generated Responses vs. \\
Groq Model and NANOGPT Outputs
\centering
\small{}
\noindent\rule{\textwidth}{0.6pt} % First line
\vspace{1pt}
\raggedright
\textbf{Query 8:} What regulatory frameworks are needed to govern the safety and ethical use of nanotechnology in consumer products?
\vspace{1pt}

\rule{\linewidth}{0.7mm}
\vspace{1pt}    
\textbf{Human Generated Answer:} \\
\begin{singlespace}
\justifying{As nanotechnology continues to evolve, the demand for clear guidelines and regulatory frameworks for its use has significantly increased. One of the primary challenges with nanotechnology is the difficulty in predicting its long-term impact on living organisms and the environment. This uncertainty arises because the technology is relatively new, and the properties of nanomaterials can differ greatly from those of bulk materials. For instance, it remains challenging to rigorously assess the cumulative effects of nanoparticles on biological systems over time.

In this context, the first crucial step is reaching a consensus on the need for a regulatory framework, acknowledging that the current level of scientific understanding is insufficient to answer many critical questions. This awareness must be cultivated not only within the scientific and engineering communities but also among the general public. Public education about the potential risks and benefits of nanotechnology is essential to foster informed discussions.

Once this common agreement is established, the next step is to develop a robust risk assessment methodology. This involves conducting more comprehensive studies on the long-term effects of nanomaterials on human health, ecosystems, and the environment—areas that are often overlooked. A comprehensive regulatory policy must address all aspects of nanomaterial use, including product development, manufacturing processes, storage, and disposal, to ensure the responsible and safe advancement of the technology.
}
\end{singlespace}

\rule{\linewidth}{0.7mm}

\vspace{1pt}
\textbf{Groq Model Answer:} \\
\begin{singlespace}
\justifying{To address the concerns surrounding nanotechnology in consumer products, regulatory frameworks should be established to ensure safety and ethical use. Some key considerations include:
Risk assessment and testing: Standardized testing protocols should be developed to assess the potential risks of nanomaterials in consumer products.
Labeling and transparency: Manufacturers should be required to label products containing nanomaterials, and provide information on their potential risks and benefits.
Toxicity and environmental impact: Regulations should be put in place to prevent the release of nanomaterials into the environment, and to mitigate their potential toxicity.
Public education and awareness: Governments and industry leaders should work together to educate consumers about the potential risks and benefits of nanotechnology in consumer products.
International cooperation: Global regulatory frameworks should be developed to ensure consistency and coordination across countries.
By establishing these regulatory frameworks, we can ensure the safe and responsible development and use of nanotechnology in consumer products.
}

\end{singlespace}

\rule{\linewidth}{0.7mm}
\vspace{1pt}    
\textbf{NANOGPT Model Answer:} \\
\begin{singlespace}
\justifying{The regulatory frameworks needed to govern the safety and ethical use of nanotechnology in consumer products are intricate and multifaceted, requiring a coordinated effort from regulatory agencies, industry, and civil society. A foundational framework is the European Union’s REACh (Registration, Evaluation, Authorization, and Restriction of Chemicals) regulation, which mandates that manufacturers provide detailed information about the chemical composition and potential risks of nanomaterials. However, REACh alone is insufficient; more specific and nuanced regulations are necessary to address the unique properties of nanomaterials, such as their small size, enhanced transport, and potential toxicity.

To enhance governance, additional measures are needed, including:

Labeling Requirements: Clear and mandatory labeling to inform consumers about the presence of nanomaterials, their risks, and potential benefits, promoting transparency and informed decision-making.

Risk Assessment and Management: Comprehensive risk assessment and management strategies tailored to the unique challenges posed by nanomaterials, ensuring their safe and responsible use.

Safety Testing: Rigorous safety testing protocols to evaluate the impact of nanomaterials on human health and the environment before market approval.

Ethical Considerations: Ethical principles, including autonomy, beneficence, non-malfeasance, and justice, should guide the development and use of nanotechnology. An Ethical Matrix analysis can help address issues like unforeseen risks, fair risk distribution, and protection for vulnerable populations and ecosystems.

Public Engagement and Education: Mechanisms for public engagement and education are essential to build trust, empower consumers, and ensure that society understands the implications of nanotechnology.

Transparency and Communication: Open communication of risks and benefits is vital to allow free and informed choice, balancing societal benefits with potential risks.

Global initiatives such as Euro-NANOTOX, BioNanoNet, and the OECD Database on Research into the Safety of Manufactured Nanomaterials exemplify efforts to standardize risk assessments and enhance research into nanotechnology safety. These frameworks must also prioritize balancing the benefits and risks of nanotechnology, ensuring that steps are taken to minimize harm to end-users, the environment, and future generations.

In sum, the regulatory frameworks for nanotechnology require adaptive, interdisciplinary approaches, integrating scientific research, ethical oversight, public involvement, and international cooperation to ensure that nanomaterials are used safely and responsibly.

References:

[1] D. Coles, L.J. Frewer (2013). Nanotechnology applied to European food production – A review of ethical and regulatory issues. DOI: 10.1016/j.tifs.2013.08.006

[2] I. Allon, N. Levine, I. Baanante (2019). Building the European Nanomedicine Research and Innovation Area: 10 years funding innovative research projects. DOI: 10.33218/PRNANO2(1).190404.1

[3] K. Juganson, A. Ivask, I. Blinova, M. Mortimer, A. Kahru (2015). NanoE-Tox: New and in-depth database concerning ecotoxicity of nanomaterials. DOI: 10.3762/bjnano.6.183
}\\
\rule{\linewidth}{0.7mm}
\noindent\rule{\textwidth}{0.4pt} % Full-width thin line

\noindent\textbf{Author Comments:}\\
\noindent\textbf{Depth of Information:}

NANOGPT offers an in-depth analysis, citing specific frameworks like the EU’s REACh regulation and international initiatives such as Euro-NANOTOX and BioNanoNet, while also discussing ethical principles and transparency measures.
Groq focuses on general principles without mentioning specific regulatory frameworks or established efforts.\\
\noindent\textbf{Technical Focus:}

NANOGPT goes beyond general regulation by incorporating advanced governance strategies, ethical considerations (e.g., ethical matrix analysis), and societal implications, making it more suitable for experts or policymakers.
Groq emphasizes basic regulatory needs without delving into nuanced challenges or ethical aspects.\\
\noindent\textbf{Structure and Clarity:}

Groq is well-structured, with clear points that are easy to follow, making it more accessible to general readers.
NANOGPT is dense and detailed, which can overwhelm non-technical audiences despite its comprehensive content.\\
\noindent\textbf{Applications and Examples:}

NANOGPT provides examples of ongoing international efforts and frameworks for nanotechnology safety, adding context and credibility.
Groq lacks specific examples or references, limiting its applicability to practical or academic discussions.\\
\noindent\textbf{References:}

NANOGPT includes multiple references, reinforcing its credibility and offering avenues for further exploration.
Groq does not provide citations, reducing its utility for research or in-depth analysis.\\
\noindent\textbf{Forward-Looking Insights:}

NANOGPT discusses the need for interdisciplinary, adaptive approaches and highlights global initiatives for advancing nanotechnology safety.
Groq focuses on establishing basic frameworks but does not address long-term or interdisciplinary strategies.\\
\noindent\textbf{Summary:}

Groq Model: Clear and concise, ideal for a general audience seeking a high-level understanding of nanotechnology regulations.
NANOGPT Model: Detailed and comprehensive, addressing frameworks, ethical considerations, and global initiatives, making it more suitable for policymakers, researchers, and experts.\\

\rule{\linewidth}{0.7mm}
\end{singlespace}

\newpage

\centering
\textbf{Table 9:} \normalsize Comparison of Biomedical query in Nanotechnology: Human-Generated Responses vs.Groq Model and NANOGPT Outputs
\centering
\small{}
\noindent\rule{\textwidth}{0.6pt} % First line
\vspace{1pt}
\raggedright
\textbf{Query 9:} How do surface functionalization techniques influence the stability and biocompatibility of nanomaterials for biomedical applications?

\vspace{1pt}

\rule{\linewidth}{0.7mm}
\vspace{1pt}    
\textbf{Human Generated Answer:} \\
\begin{singlespace}
\justifying{Nanomaterials have the unique advantage of exhibiting surface chemistry distinct from traditional bulk materials, offering opportunities for functionalization with various chemical groups. This ability to modify their surface plays a crucial role in enhancing their interactions with biological systems. Surface functionalization techniques are particularly important in improving biocompatibility, as they can reduce the inherent toxicity of certain nanomaterials, increase cellular uptake, and enable targeted drug or molecule delivery to specific sites in the body.

By attaching polymers, biomolecules, or specific ligands to the surface of nanomaterials, their interactions with biological environments can be finely tuned. For instance, functionalization with biocompatible polymers like PEG (polyethylene glycol) can help shield nanomaterials from recognition by the immune system, minimizing immune responses and promoting longer circulation times in the bloodstream. Additionally, conjugating nanomaterials with targeting ligands, such as antibodies or peptides, allows them to selectively bind to specific cells or tissues, increasing the precision of drug delivery, particularly in cancer therapy or gene delivery.

These modifications not only improve the therapeutic efficacy but also reduce potential side effects by ensuring that nanomaterials remain active for extended periods in the body while minimizing off-target interactions. This enhanced control over the interaction between nanomaterials and their biological environment demonstrates the powerful potential of surface functionalization to make nanotechnology more effective and safer in biomedical applications.
}
\end{singlespace}

\rule{\linewidth}{0.7mm}

\vspace{1pt}
\textbf{Groq Model Answer:} \\
\begin{singlespace}

\justifying{Surface functionalization is a crucial step in modifying the properties of nanomaterials for biomedical applications. It involves attaching specific molecules or groups to the surface of nanomaterials to enhance their stability, biocompatibility, and bioactivity.
Techniques used for surface functionalization include:
Chemical conjugation: Covalent bonding of molecules to the surface of nanomaterials.
Physical absorption: Non-covalent attachment of molecules to the surface of nanomaterials.
Bioconjugation: Attachment of biomolecules, such as proteins or peptides, to the surface of nanomaterials.
Surface functionalization can improve the:
Stability of nanomaterials by reducing aggregation and aggregation-related toxicity.
Biocompatibility by reducing immune responses and improving cell compatibility.
Bioactivity by enhancing the ability of nanomaterials to interact with cells and tissues.
Overall, surface functionalization is a powerful tool for tailoring the properties of nanomaterials for specific biomedical applications.
}
\end{singlespace}

\rule{\linewidth}{0.7mm}
\vspace{1pt}    
\textbf{NANOGPT Model Answer:} \\
\begin{singlespace}
\justifying{Surface functionalization techniques play a crucial role in influencing the stability and biocompatibility of nano materials for biomedical applications. By modifying the surface properties of nano materials, surface functionalization techniques can alter their interactions with biological systems, leading to improved stability and biocompatibility.

For example, surface functionalization techniques such as plasma treatment, chemical modification, and bioconjugation can be used to modify the surface properties of nano materials, such as their hydrophilicity, hydrophobicity, and charge. These modifications can influence the interactions between the nano materials and biological systems, such as cells and tissues, and can affect their stability and biocompatibility.

In particular, surface functionalization techniques can be used to:

Improve the biocompatibility of nano materials by reducing their toxicity and improving their interactions with biological systems.
Enhance the stability of nano materials by reducing their aggregation and improving their dispersion in biological fluids.
Target specific biological systems or cells by modifying the surface properties of nano materials to interact with specific receptors or molecules.
Overall, surface functionalization techniques are an important tool for improving the stability and biocompatibility of nano materials for biomedical applications, and can help to ensure their safe and effective use in medical treatments.

References:

[1] Małgorzata Jurak, Agnieszka Ewa Wiącek, Agata Ładniak, Kacper Przykaza, Klaudia Szafran (2021). What affects the biocompatibility of polymers?. DOI: 10.1016/j.cis.2021.102451

[2] P Slepicka, NS Kasalkova, J Siegel, Z Kolska, L Bacakova, V Svorcik (2015). Nano-structured and functionalized surfaces for cytocompatibility improvement and bactericidal action. DOI: 10.1016/
j.biotechadv.2015.01.001

[3] PN Navya, HK Daima (2016). Rational engineering of physicochemical properties of nanomaterials for biomedical applications with nanotoxicological perspectives. DOI: 10.1186/s40580-016-0064-z}\\
\rule{\linewidth}{0.7mm}
\noindent\rule{\textwidth}{0.4pt}\\ % Full-width thin line

\noindent\textbf{Author Comments:}\\
\noindent\textbf{Depth of Information:}

NANOGPT offers a more detailed explanation, discussing additional techniques like plasma treatment and their influence on properties such as hydrophilicity, hydrophobicity, and charge. It also explains how functionalization can be tailored to target specific biological receptors or molecules.
Groq focuses on general improvements (e.g., stability, biocompatibility) without discussing the underlying mechanisms or additional techniques.\\
\noindent\textbf{Technical Focus:}

NANOGPT provides richer insights into how surface functionalization affects interactions with biological systems and addresses toxicity and dispersion challenges in biological fluids.
Groq is less technical, making it more accessible to general readers but less informative for researchers or practitioners.\\
\noindent\textbf{Structure and Clarity:}

Groq is clear and well-structured, listing techniques and their benefits in a straightforward manner.
NANOGPT is denser and more comprehensive, which may require a stronger technical background to fully understand.\\
\noindent\textbf{Applications and Specificity:}

NANOGPT discusses how surface functionalization can be used to target specific cells or receptors, offering greater specificity for biomedical applications.
Groq mentions general benefits but does not link these to specific biomedical use cases.\\
\noindent\textbf{References:}

NANOGPT includes multiple references, enhancing its credibility and offering sources for further exploration.
Groq does not provide citations, limiting its utility for academic or research purposes.\\
\noindent\textbf{Forward-Looking Insights:}

NANOGPT emphasizes the role of surface functionalization in ensuring safe and effective use of nanomaterials in medical treatments, addressing long-term considerations like nanotoxicology.
Groq focuses on current benefits without discussing future implications or challenges.\\
\noindent\textbf{Summary:}

Groq Model: Clear and structured, providing a basic overview of surface functionalization techniques and their benefits, suitable for non-technical audiences.
NANOGPT Model: More detailed and technically comprehensive, discussing additional techniques, specific interactions, and challenges, making it ideal for researchers or professionals in biomedical nanotechnology.

\end{singlespace}
\rule{\linewidth}{0.7mm}

\newpage

\centering
\textbf{Table 10:} \normalsize Comparison of Desalination in Nanotechnology: Human-Generated Responses vs. \\
Groq Model and NANOGPT Outputs
\centering
\small{}
\noindent\rule{\textwidth}{0.6pt} % First line
\vspace{1pt}
\raggedright
\textbf{Query 10:} What are the applications of nanotechnology in desalination?

\vspace{1pt}
\rule{\linewidth}{0.7mm}
\vspace{1pt}    
\textbf{Human Generated Answer:} \\
\begin{singlespace}
\justifying{Water desalination is an excellent use case for nanomaterials. Nanomaterials such as metal oxide nanoparticles, graphene, and carbon nanotubes (CNTs) are effective materials for desalination processes. These materials offer a high surface area and strong adsorption properties, allowing for the efficient removal of salt ions from seawater through adsorption. The morphology and chemical functionality of nanomaterials are crucial in determining their performance during desalination, depending on the specific desalination method being used.
Key factors such as pore size and surface chemistry must be carefully considered to optimize adsorption properties, where a high surface-to-volume ratio is particularly advantageous. In terms of chemical properties, the catalytic activity of nanomaterials significantly influences the efficiency of processes like electrodialysis and ion exchange. For membrane-based desalination methods, considerations like pore size and surface charge are critical for achieving high salt rejection rates while maintaining the ability to handle large volumes of water.
Additionally, the chemical and physical stability of nanomaterials, as well as their potential environmental impact, must be thoroughly evaluated. Ensuring that nanomaterials are both effective and safe in desalination applications is essential for their long-term use in sustainable water purification technologies.
Nano adsorbents, like activated carbon nanoparticles and metal oxides, act as efficient sponges, capturing contaminants like heavy metals, organic pollutants, and even bacteria from water. This additional purification step ensures the desalinated water meets rigorous quality standards. 2D nanomaterials like graphene, maxene, MoS2, also are good candidates for high efficiency desalination.
}
\end{singlespace}

\rule{\linewidth}{0.7mm}

\vspace{1pt}
\textbf{Groq Model Answer:} \\
\begin{singlespace}
\justifying{Nanotechnology has been increasingly utilized in desalination processes to enhance efficiency, reduce energy consumption, and mitigate environmental impacts. Some of the key applications of nanotechnology in desalination include:
Membrane technology: Nanomaterials are used to create ultrafiltration and reverse osmosis membranes, which can effectively remove salt and other impurities from seawater.
Antifouling coatings: Nanocoatings can be applied to membranes to prevent fouling by bacteria, algae, and other microorganisms, reducing the need for frequent cleaning and maintenance.
Nanostructured surfaces: The incorporation of nanostructured surfaces can improve the flow rates and permeability of membranes, leading to increased desalination efficiency.
Nanofiltration: Nanofiltration membranes can be used to remove dissolved solids and other impurities from seawater, producing fresh water suitable for drinking and other applications.
Overall, the integration of nanotechnology in desalination processes has the potential to increase efficiency, reduce costs, and provide a more sustainable solution for water scarcity issues.
}
\end{singlespace}

\rule{\linewidth}{0.7mm}
\vspace{1pt}    
\textbf{NANOGPT Model Answer:} \\
\begin{singlespace}
\justifying{Nanotechnology plays a pivotal role in advancing desalination processes by offering innovative solutions to improve the efficiency and effectiveness of water treatment. Its applications in desalination are diverse and impactful. One significant application is the enhancement of nanofiltration and reverse osmosis membranes, which facilitates the efficient removal of impurities and contaminants. The incorporation of multi-walled carbon nanotubes (MWCNTs) in ion exchange and reverse osmosis membranes has been particularly effective in increasing desalination efficiency without compromising water conductance, making it possible to produce fresh water from seawater and brackish water for human consumption.

In addition to improving conventional membrane technologies, nanotechnology has also driven the development of advanced methods such as forward osmosis and membrane distillation. These technologies leverage the unique properties of nanomaterials to achieve more effective treatment of seawater and brackish water, showing great promise in expanding the capabilities of desalination.

Nanocomposites play a vital role in enhancing the performance of desalination membranes by improving ionic permeability and ion selectivity, reducing energy consumption and operational costs. Nanomaterials such as carbon nanotubes and other advanced composites not only improve membrane efficiency but also enhance their durability and resistance to fouling, ensuring longer operational life and reliability.

Furthermore, nanotechnology has facilitated the development of supercapacitor-based water filters, which offer highly efficient removal of salts and other contaminants from seawater. Another exciting innovation involves nanocomposite-based solar desalination systems, which utilize solar energy to provide sustainable and energy-efficient solutions for water treatment.

Overall, the integration of nanotechnology into desalination processes has the potential to revolutionize water treatment by significantly improving efficiency, sustainability, and accessibility, addressing the growing global demand for fresh water.

References:

[1]Rafael Castaneda, Edgar Zayago Lau, Eduardo Robles-Belmont, Sein Leon Silva (2017). Review of nanotechnology value chain for water treatment applications in Mexico. DOI:\\ 
10.18799/24056529/2017/13/113

[2] Muktar Nono Mohammed, MA Zaed, R. Saidur, K.H. Tan (2024). Nanocomposite-based solar desalination: Recent developments and future prospects. DOI: 10.1016/j.jwpe.2024.105733

[3] A Zapata-Sierra, M Cascajares, A Alcayde, F Manzano-Agugliaro (2021). Worldwide research trends on desalination. DOI: 10.1016/j.desal.2021.115305}\\
\rule{\linewidth}{0.7mm}
\noindent\rule{\textwidth}{0.4pt}\\ % Full-width thin line

\noindent\textbf{Author Comments:}\\
\noindent\textbf{Depth of Information:}\\

NANOGPT offers a more comprehensive discussion, detailing specific nanomaterials like multi-walled carbon nanotubes (MWCNTs) and nanocomposites, and their roles in improving membrane efficiency, durability, and fouling resistance.
Groq focuses on general applications without delving into the materials or mechanisms behind these innovations.\\
\noindent\textbf{Technical Focus:}\\

NANOGPT introduces advanced desalination methods such as forward osmosis, membrane distillation, and solar desalination using nanocomposites, demonstrating its technical focus and future-oriented perspective.
Groq emphasizes traditional methods like reverse osmosis and ultrafiltration without exploring innovative technologies.\\
\noindent\textbf{Structure and Clarity:}\\

Groq is well-organized with simple, categorized points, making it easier to follow for general audiences.
NANOGPT is denser and more technical, requiring a stronger background in the subject to fully comprehend.\\
Applications and Specificity:\\

NANOGPT provides specific examples, such as MWCNT-enhanced membranes and supercapacitor-based water filters, adding detail and context to its explanations.
Groq discusses applications in general terms without linking them to particular nanomaterials or innovations.\\
\noindent\textbf{References:}\\

NANOGPT includes multiple references, lending credibility and allowing readers to explore further research.
Groq does not provide citations, limiting its utility for academic or professional purposes.\\

\noindent\textbf{Forward-Looking Insights:}\\

NANOGPT highlights future prospects like solar desalination and addresses the sustainability of nanotechnology in water treatment.
Groq focuses on current benefits without addressing future advancements or sustainability concerns.\\

\noindent\textbf{Summary:}\\
 
Groq Model: Clear and accessible, providing a basic overview of nanotechnology’s role in desalination, suitable for general audiences.
NANOGPT Model: Detailed and comprehensive, with specific examples, emerging technologies, and references, making it ideal for researchers, professionals, and technical readers.

\rule{\linewidth}{0.7mm}

\end{singlespace}
\newpage
\textbf{Table 11:} \normalsize Comparison of Carbide Formation: Human-Generated Responses vs. \\
Groq Model and NANOGPT Outputs
\centering
\small{}
\noindent\rule{\textwidth}{0.6pt} % First line
\vspace{1pt}
\raggedright
\textbf{Query 11:} For the formation of carbides and nitrides of Ti, V, Mo, and W, describe the changes that occur to the parent metal during the intercalation of carbon into the lattice of the parent metal to form metal carbide. 
?

\vspace{1pt}
\rule{\linewidth}{0.7mm}
\vspace{1pt}    
\textbf{Human Generated Answer:} \\
\begin{singlespace}
\justifying{
Transition metal carbides or nitrides are formed by the incorporation of carbon or nitrogen atom, respectively, into the interstitial transition metal network. The formation of metal carbides and nitrides alters the nature of the d-band of the parent transition metal, leading to improved catalytic properties relative to those of the parent transition metals and making them similar to group VIII noble metals. The formation of interstitial compounds results in the metal lattice expansion, increasing the metal-metal distance and causing a contraction of the metal d-band. This contraction in the d-band results in a higher density of states (DoS) near the Fermi level, despite the charge transfer from the metal to the carbon or nitrogen atoms. The formation of M-C and M-N bonds leads to significant redistributions of the DoS below and above the Fermi level, resembling that of group VIII noble metals.

}
\end{singlespace}

\rule{\linewidth}{0.7mm}

\vspace{1pt}
\textbf{Groq Model Answer:} \\
\begin{singlespace}
\justifying{The formation of carbides from transition metals such as Ti, V, Mo, and W involves significant changes to their crystal structure and electronic configuration. As carbon atoms intercalate into the lattice of the parent metal, they replace some of the metal atoms, resulting in a distorted lattice arrangement. The addition of carbon atoms also leads to a decrease in the metal's crystal lattice parameter. Furthermore, the electronegativity of carbon tends to increase the ionic character of the bond with the metal atom, causing a shift towards a more BCC or simple cubic structure. This distortion often increases the melting point and hardness of the resulting metal carbide.
}
\end{singlespace}

\rule{\linewidth}{0.7mm}
\vspace{1pt}    
\textbf{NANOGPT Model Answer:} \\
\begin{singlespace}
\justifying{The intercalation of carbon into the lattice of parent metals such as titanium (Ti), vanadium (V), molybdenum (Mo), and tungsten (W) to form metal carbides involves several significant changes. Initially, the parent metal's crystal structure is altered as carbon atoms are introduced into the lattice. This process can lead to the expansion of the lattice parameters due to the larger size of carbon compared to the metal atoms.

As carbon intercalates, it can also influence the electronic structure of the metal, potentially enhancing its conductivity and altering its bonding characteristics. The presence of carbon can create new bonding interactions, resulting in a more complex electronic environment. This change can enhance the hardness and strength of the resulting carbide compared to the parent metal.

Additionally, the mechanical properties of the metal are modified, often leading to increased brittleness due to the formation of a more rigid structure. The resulting metal carbide typically exhibits high melting points and exceptional hardness, making it suitable for various applications, including cutting tools and wear-resistant coatings. Overall, the transformation from metal to carbide involves significant structural, electronic, and mechanical changes that enhance the material's performance characteristics.

References:

[1] Cristina Giordano, Markus Antonietti (2011). Synthesis of crystalline metal nitride and metal carbide nanostructures by sol–gel chemistry. DOI: 10.1016/j.nantod.2011.06.002

[2] Oladipo Folorunso, Neeraj Kumar, Yskandar Hamam, Rotimi Sadiku, Suprakas Sinha Ray (2021). Recent progress on 2D metal carbide/nitride (MXene) nanocomposites for lithium-based batteries. DOI: 10.1016/j.flatc.2021.100281}\\
\rule{\linewidth}{0.7mm}

\noindent\textbf{Author Comments:}\\
\noindent\textbf{Depth of Information}\\
Groq Model:

%The Groq response offers a general description of carbide formation, focusing on lattice distortion, electronegativity changes, and basic mechanical properties such as increased hardness and melting point. However, it lacks specific technical details or deeper exploration of structural or electronic changes.
The Groq response includes some of the key terms such as intercalation, distortion, electronegativity, and lattice to describe carbides formation from the parent metals. However, the description of these terms in relation to the carbide formation are basis as it does not include details or deeper exploration of how the structural or electronic changes influence material properties.

NANOGPT Model:
NANOGPT provides more detailed and organized information of the carbide formation. It indicated the impact of carbon intercalation on lattice parameters, electronic structure, and bonding characteristics. It goes further to describe the effects on conductivity, mechanical properties, and specific applications like cutting tools and coatings. This makes the response more comprehensive. 

\noindent\textbf{Technical Focus}\\
Groq Model:
Focuses primarily on the basic crystallographic and ionic changes, with minimal discussion on the electronic or advanced mechanical properties. It stated that  carbon replace some of the metal atoms, but its known that carbon occupy the interstitial transition metal network. It also generates wrong statement like "the addition of carbon atoms also leads to a decrease in the metal's crystal lattice parameter", whereas it increases the metal's crystal lattice parameter.

NANOGPT Model:
Delves into the interplay between lattice structure and electronic properties, highlighting advanced materials chemistry concepts such as conductivity enhancement and the complex bonding environment. It analysis is closer to human generated answer, although it would be more helpful if it provide further details about make the bonding complex.

\noindent\textbf{Structure and Clarity}\\
Groq Model:
The response is concise and organized in a straightforward manner, making it accessible to general readers. However, the simplicity sacrifices depth and technical accuracy.

NANOGPT Model:
The answer is dense and technical, requiring familiarity with materials science terminology. While well-organized, the complexity may make it less accessible to non-experts.

\noindent\textbf{Applications and Specificity}\\
Groq Model:
Mentions broad applications like increased hardness and melting point but does not tie these properties to specific use cases or detailed mechanisms.

NANOGPT Model:
Discusses specific applications such as cutting tools and wear-resistant coatings. It also links structural and electronic changes to these applications, making the explanation contextually rich.

\noindent\textbf{Forward-Looking Insights}\\
Groq Model:
Focuses on existing properties without exploring future implications or advancements in carbide materials.

NANOGPT Model:
Highlights advancements and applications in cutting-edge fields like high-performance coatings, suggesting a forward-looking perspective.

\rule{\linewidth}{0.7mm}

\end{singlespace}
\newpage

\centering
\textbf{Table 12:} \normalsize Comparison of Helmholtz layer charge storage: Human-Generated Responses vs. \\
Groq Model and NANOGPT Outputs
\centering
\small{}
\noindent\rule{\textwidth}{0.6pt} % First line
\vspace{1pt}
\raggedright
\textbf{Query 12:} Describe how charges are store in the double layer or Helmholtz layer of a micropore electrode.

\vspace{1pt}
\rule{\linewidth}{0.7mm}
\vspace{1pt}    
\textbf{Human Generated Answer:} \\
\begin{singlespace}
\justifying{In an electrolyte solution, ions are typically surrounded by a "shell" of solvent molecules, a state known as solvation. These solvated ions move freely in the bulk of the electrolyte. When an external electric field is applied, such as when a capacitor or battery is charged, the ions are drawn to the electrode surface, particularly into the micropores (pores smaller than 2 nm). These micropores provide a large surface area capable of storing significant charge. As ions enter these narrow micropores, the confined space often causes some solvent molecules to separate from the ions, a process called partial desolvation. This reduces the ions' effective size, enabling them to fit into the small pores and come closer to the electrode surface. Partially desolvated ions form stronger interactions with the electrode, enhancing charge storage. Inside the micropores, the ions align themselves near the electrode surface, forming a layer of charge known as the inner layer or Stern layer. A second, oppositely charged layer forms in the electrolyte adjacent to this inner layer, creating an electrical double layer. This double layer acts like a parallel plate capacitor, storing energy as an electrostatic potential difference.
}
\end{singlespace}

\rule{\linewidth}{0.7mm}

\vspace{1pt}
\textbf{Groq Model Answer:} \\
\begin{singlespace}
\justifying{The double layer or Helmholtz layer in a micropore electrode is a region where charges are stored due to the capacitance effects present at the electrode-solution interface. This is typically observed in the nanoscale or microscale pores in electrodes, where the pore diameter is comparable to or smaller than the diameter of hydrated ions.

In the double layer, charges are stored in two ways. Firstly, ions that are attracted to the electrode surface through electrostatic forces get confined to a thin layer near the surface due to the strong interfacial potentials. These ions contribute to the surface capacitance by storing charge. Secondly, the electrodes themselves can be positively or negatively charged depending on the properties of the material. When an electric potential is applied across the electrode, the electrodes can store additional charge, leading to variations in surface charge density.

The charge storage in the double layer is influenced by several factors such as electrolyte concentration, electrode material, potential, surface roughness and pore sizes. This double layer play a crucial role in various electrochemical processes.
}
\end{singlespace}

\rule{\linewidth}{0.7mm}
\vspace{1pt}    
\textbf{NANOGPT Model Answer:} \\
\begin{singlespace}
\justifying{In the double layer or Helmholtz layer of a micropore electrode, charges are stored through the accumulation of ions at the interface between the electrode and the electrolyte solution. When a potential difference is applied, cations from the solution are attracted to the negatively charged electrode surface, while anions are repelled. This creates a region where positive and negative charges are separated, forming an electrical double layer. The arrangement of these charges is influenced by factors such as the concentration of ions in the solution, the surface charge density of the electrode, and the geometry of the micropores. The resulting potential difference across the double layer allows for the storage of electrical energy, which can be utilized in applications like capacitive deionization.

References:

[1] A.V. Delgado, M.L. Jiménez, G.R. Iglesias, S. Ahualli (2019). Electrical double layers as ion reservoirs: applications to the deionization of solutions. DOI: 10.1016/j.cocis.2019.09.003

[2] Ángel V. Delgado, Silvia Ahualli, María M. Fernández, Guillermo R. Iglesias, María L. Jiménez (2018). Charge and Potential Distribution in the Electrical Double Layer of Porous Materials: Models. DOI: 10.1016/b978-0-12-811370-7.00001-2}

\rule{\linewidth}{0.7mm}

\noindent\textbf{Author Comments:}\\
\noindent\textbf{Depth of Information}\\

Groq Model: The response provides a general overview of charge storage in the double layer of a micropore electrode. It explains basic concepts such as capacitance effects at the electrode-solution interface and mentions factors influencing charge storage like electrolyte concentration and pore sizes. However, it lacks detailed explanations of the underlying mechanisms and does not delve into specific models or theoretical frameworks that describe the double layer's behavior in microporous electrodes.

NANOGPT Model: This answer offers a more detailed analysis of how charges are stored in the double layer. It describes the accumulation of ions at the electrode-electrolyte interface and explains the separation of charges forming the electrical double layer. The response discusses the influence of ion concentration, surface charge density, and micropore geometry on charge arrangement. By mentioning applications like capacitive deionization and providing references, it adds depth and allows readers to explore the topic further.\\

\noindent\textbf{Technical Focus}\\

Groq Model: The focus is on basic electrostatic interactions and general factors affecting charge storage. It touches upon the role of ions in surface capacitance and mentions variables like electrolyte concentration and pore size. However, it does not engage with advanced electrochemical theories or models related to the double layer in microporous materials.

NANOGPT Model: The answer goes into the electrochemical processes at the microscopic level, explaining how cations are attracted to the negatively charged electrode surface while anions are repelled, leading to charge separation. It highlights factors influencing the electrical double layer's formation and directly connects these technical aspects to practical applications like capacitive deionization, demonstrating a strong technical focus.

\noindent\textbf{Structure and Clarity}\\

Groq Model: The response is concise and organized logically, making it accessible to readers with a basic understanding of electrochemistry. It introduces concepts in a straightforward manner but may be too simplistic for those seeking a deeper technical understanding.

NANOGPT Model: The answer is well-structured and presents information in a clear, logical sequence. It uses technical terminology appropriately and explains complex concepts effectively. Inclusion of references enhances clarity by providing avenues for readers to obtain more detailed information, although the density of information might be challenging for non-experts.

\noindent\textbf{Applications and Specificity}\\

Groq Model: While it mentions that the double layer plays a crucial role in various electrochemical processes, it does not specify any particular applications or link the described properties to specific technologies or mechanisms.

NANOGPT Model: The response explicitly mentions capacitive deionization as an application of charge storage in the electrical double layer. By doing so, it provides context and demonstrates how the discussed phenomena are relevant in practical, real-world scenarios. The inclusion of specific applications adds significant value to the explanation.

\noindent\textbf{Forward-Looking Insights}\\

Groq Model: The answer focuses on existing knowledge and does not provide insights into future implications, advancements, or emerging research areas related to the double layer in micropore electrodes.

NANOGPT Model: While primarily explanatory, the inclusion of recent references suggests an awareness of ongoing research in the field. However, it does not specifically offer forward-looking perspectives or discuss potential future developments or innovations stemming from the current understanding of the double layer.

\end{singlespace}
\newpage
\centering
\textbf{Table 13:} \normalsize Comparison of responses for query on Adsorption: Human-Generated Responses vs. \\
Groq Model and NANOGPT Outputs
\centering
\small{}
\noindent\rule{\textwidth}{0.6pt} % First line
\vspace{1pt}
\raggedright
\textbf{Query 13:} According to Sabatier Principle, can I achieve optimal performance when the adsorbate binds tightly to the catalyst surface.

\vspace{1pt}
\rule{\linewidth}{0.7mm}
\vspace{1pt}    
\textbf{Human Generated Answer:} \\
\begin{singlespace}
\justifying{The Sabatier Principle is a concept in catalysis that describes the relationship between the strength of interaction (or binding energy) between a catalyst and the reactants. The Sabatier Principle suggests that there is an optimal binding energy where the catalyst binds reactants strongly enough to promote reaction but not so strongly that it hinders the release of products and not too weak that it may not hold onto them long enough to lower the activation energy for the reaction to proceed. This optimal interaction maximizes the catalytic activity, as it allows the reactants to adsorb, react, and then desorb efficiently.
}
\end{singlespace}

\rule{\linewidth}{0.7mm}

\vspace{1pt}
\textbf{Groq Model Answer:} \\
\begin{singlespace}
\justifying{Actually, the Sabatier principle suggests the opposite. It states that optimal catalytic performance is achieved when the adsorbate binds moderately to the catalyst surface. If the adsorbate binds too tightly, it can lead to deactivation of the catalyst due to strong adsorption, making it difficult to desorb and regenerate the catalyst. On the other hand, if the adsorbate binds too weakly, it may not be effectively activated, resulting in poor catalytic performance. Therefore, a moderate binding energy is essential for achieving optimal catalytic performance.
}
\end{singlespace}

\rule{\linewidth}{0.7mm}
\vspace{1pt}    
\textbf{NANOGPT Model Answer:} \\
\begin{singlespace}
\justifying{According to the Sabatier principle, optimal performance is achieved when the adsorbate binds neither too tightly nor too weakly to the catalyst surface. If the adsorbate binds too tightly, it may become difficult for the reactants to access the active sites, leading to a low overall reaction rate. Conversely, if the binding is too weak, the adsorbate may not stay on the surface long enough to react. Therefore, there is a balance that must be struck for optimal catalytic activity.

References:

[1] Anshuman Goswami, Hanyu Ma, William F. Schneider (2022). Consequences of adsorbate-adsorbate interactions for apparent kinetics of surface catalytic reactions. DOI: 10.1016/j.jcat.2021.12.005

[2] Sarah M. Stratton, Shengjie Zhang, Matthew M. Montemore (2023). Addressing complexity in catalyst design: From volcanos and scaling to more sophisticated design strategies. DOI: 10.1016/j.surfrep.2023.100597\\

}

\noindent\textbf{Author Comments:}\\

\noindent\textbf{Depth of Information}\\

Groq Model: The Groq response provides a basic explanation of the Sabatier principle, emphasizing the necessity of moderate binding energy for optimal catalytic performance. It outlines the consequences of adsorbates binding too tightly or too weakly but doesn't delve into detailed mechanisms or provide specific examples.

NANOGPT Model: The NANOGPT answer offers a more comprehensive analysis. It discusses the balance required in adsorbate binding, explaining how tight binding can hinder reactant access to active sites and how weak binding can result in insufficient interaction time. The inclusion of recent references adds depth and shows engagement with current research.

\noindent\textbf{Technical Focus}\\

Groq Model: Focuses on the general concept of the Sabatier principle without exploring underlying kinetic or thermodynamic aspects. The explanation remains on a fundamental level, avoiding complex technical details.

NANOGPT Model: Goes into technical aspects by linking binding strength to reaction rates and catalyst activity. References to active sites and the dynamics of adsorbate interaction indicate a stronger technical emphasis, appealing to readers with a background in catalysis or surface chemistry.

\noindent\textbf{Structure and Clarity}\\

Groq Model: The response is clearly structured and easy to understand, making it accessible to a broad audience. The logical flow aids in comprehension, but the simplicity may limit the depth of understanding for advanced readers.

NANOGPT Model: While the answer is well-organized, it incorporates more technical terminology, which might challenge readers unfamiliar with the subject matter. The inclusion of references though, is academically appropriate.

\noindent\textbf{Applications and Specificity}\\

Groq Model: Discusses the implications of binding strength in a general sense but lacks specific examples or real-world applications of the Sabatier principle in catalyst design or industrial processes.

NANOGPT Model: Hints at practical applications by mentioning reaction rates and the necessity of balancing binding energies for catalytic activity. The cited references suggest specific studies and advanced strategies in catalyst design.

\noindent\textbf{Forward-Looking Insights}\\

Groq Model: The answer remains focused on explaining the established principle without considering future developments or emerging research areas in catalysis.

NANOGPT Model: By citing recent studies from 2022 and 2023, the response demonstrates awareness of ongoing research and potential advancements in catalyst design strategies. This forward-looking perspective indicates an engagement with the evolving field.

\end{singlespace}
\newpage

\textbf{Table 14:} \normalsize Comparison for Pt-Electrocatalyst: Human-Generated Responses vs. \\
Groq Model and NANOGPT Outputs
\centering
\small{}
\noindent\rule{\textwidth}{0.6pt} % First line
\vspace{1pt}
\raggedright
\textbf{Query 14:} In Pt-electro-catalyst, how can we measure active electrochemical surface area?

\vspace{1pt}
\rule{\linewidth}{0.7mm}
\vspace{1pt}    
\textbf{Human Generated Answer:} \\
\begin{singlespace}
\justifying{Underdeposition of hydrogen is used to measure the surface area, such that during the forward reaction, H Is deposited(adsorbed)  (H+  +   e- -> H*) between V = 0.4 and V =0 with respect to SHE. For each H atom adsorbed on Pt, one electron is removed, so the measured current can be mapped directly on the number of H atoms deposited. Since the size of H (how much surface area it occupies on Pt) is known, then surface area of Pt will be determined.}
\end{singlespace}

\rule{\linewidth}{0.7mm}

\vspace{1pt}
\textbf{Groq Model Answer:} \\
\begin{singlespace}
\justifying{Measuring active electrochemical surface area (ECSA) in Pt-electro-catalysts is crucial for understanding their performance. One common method is the cyclic voltammetry (CV) technique. By scanning the potential between 0.05 and 1.2 V vs. RHE (reversible hydrogen electrode), we can observe the hydrogen adsorption/desorption peaks, which are directly proportional to the ECSA. The charge under these peaks is then calculated using the hydrogen adsorption/desorption charge (typically 210 μC/cm²) and the Pt loading. This method provides a reliable estimate of the ECSA, allowing us to optimize the catalyst's performance and durability.}

\end{singlespace}

\rule{\linewidth}{0.7mm}
\vspace{1pt}    
\textbf{NANOGPT Model Answer:} \\
\begin{singlespace}
\justifying{Active electrochemical surface area (ECSA) in Pt-electro-catalysts can be measured using two primary methods. One approach is the non-Faradaic double-layer capacitance (CDL) method, which approximates ECSA by analyzing the capacitive current in relation to the scan rate and specific capacitance of the material. The other method involves surface faradaic reactions, such as hydrogen underpotential deposition (H-UPD) or CO stripping, which can directly estimate ECSA based on the charge associated with these reactions. Additionally, morphological characterization techniques like electron microscopy and surface area measurement methods such as Brunauer-Emmett-Teller (BET) can also be utilized as alternative approaches to assess ECSA.}
\end{singlespace}

References:

[1] A. Seifitokaldani, O. Savadogo (2015). Electrochemically Stable Titanium Oxy-Nitride Support for Platinum Electro-Catalyst for PEM Fuel Cell Applications. DOI: 10.1016/j.electacta.2015.03.189

[2] Mengmeng Lao, Peng Li, Yinzhu Jiang, Hongge Pan, Shi Xue Dou, Wenping Sun (2022). From fundamentals and theories to heterostructured electrocatalyst design: An in-depth understanding of alkaline hydrogen evolution reaction. DOI: 10.1016/j.nanoen.2022.107231

\noindent\textbf{Author Comments:}\\
\noindent\textbf{Depth of Information}

Groq Model:

The response offers a basic explanation of measuring ECSA using the cyclic voltammetry (CV) technique. It describes the potential range scanned, the observation of hydrogen adsorption/desorption peaks, and the calculation involving the charge (typically 210$\mu\text{C/cm}^2$) and platinum loading.

Analysis: While it covers the fundamental aspects of one common method, it lacks a discussion of alternative techniques or deeper insights into the underlying electrochemical principles.

NANOGPT Model:

This answer provides a detailed analysis of multiple methods for measuring ECSA. It discusses both non-Faradaic methods like the double-layer capacitance (CDL) and Faradaic reactions such as hydrogen underpotential deposition (H-UPD) and CO stripping. By exploring various techniques and mentioning morphological characterization methods like electron microscopy and BET, it offers a more comprehensive overview.

\noindent\textbf{Technical Focus}

Groq Model:

The focus is primarily on the CV technique related to hydrogen adsorption/desorption. It mentions specific parameters like the potential range and charge values. The technical content is accurate but limited to a single method, without exploring more advanced electrochemical concepts or other measurement techniques.

NANOGPT Model:

The response goes into advanced electrochemical methods, explaining how capacitive currents relate to scan rates and specific capacitance in the CDL method. It also covers surface Faradaic reactions and alternative morphological techniques. This displays a strong technical focus, highlighting a deeper understanding of electrochemical surface area measurements and the interplay between different characterization methods.

\noindent\textbf{Structure and Clarity}

Groq Model:

The answer is concise and easy to follow. It explains the procedure in a step-by-step manner. The simplicity aids in accessibility, but glosses over complex concepts, potentially leaving out critical technical details for expert readers.

NANOGPT Model:

While the response is information-rich, it uses dense technical language and introduces multiple concepts in quick succession. This complexity may challenge readers unfamiliar with the terminology, but it is well-organized for an audience with a background in electrochemistry or materials science.

\noindent\textbf{Applications and Specificity}

Groq Model:

Although it does not offer any particular applications or settings, the response acknowledges the significance of monitoring ECSA for maximising catalyst performance and durability. It discusses the method's applicability in a generic way without providing specific use cases or in-depth examples.

NANOGPT Model:

By mentioning methods like electron microscopy and BET analysis, the response suggests applications in cutting-edge research. The inclusion of contemporary references implies relevance to ongoing scientific research and technical advancements, even while it does not specifically link approaches to particular applications.

\noindent\textbf{Forward-Looking Insights}

Groq Model:

The focus remains on established methods without mentioning recent advancements or future directions in ECSA measurement. This limits the perspective to current practices, missing an opportunity to discuss emerging trends or innovations.

NANOGPT Model:

By citing recent studies from 2015 and 2022, the answer incorporates contemporary research findings.
Analysis: This inclusion indicates a forward-looking approach, acknowledging ongoing developments and the evolution of techniques in the field.

\end{document}